\documentclass[letterpaper, paper,10pt]{AAS}		

\usepackage{bm}
\usepackage{amsmath}
\usepackage[colorlinks=true, pdfstartview=FitV, linkcolor=black, citecolor= black, urlcolor= black]{hyperref}
\usepackage{color,soul}
\usepackage{overcite}
\usepackage{footnpag}
\usepackage{float}
\usepackage{tabularx}
\usepackage{array}  
\usepackage{makecell}
\usepackage{booktabs}
\usepackage{subcaption}
\usepackage{comment}
\usepackage{graphicx}
\usepackage{caption}
\usepackage{changepage}
\usepackage{siunitx}
\usepackage{nameref}  
\usepackage{svg}
\usepackage{multirow}
\usepackage{svg}

\PaperNumber{26-146}

\begin{document}

\title{Digital and Robotic Twinning for Validation of Proximity Operations and Formation Flying}

\author{ Zahra Ahmed\thanks{The first two authors contributed equally to the paper.} \thanks{PhD Candidate, Stanford University, Department of Aeronautics and Astronautics, Stanford, CA 94305.} , Emily Bates\footnotemark[1] \footnotemark[2], Pol Francesch Huc\footnotemark[2], Samuel Y. W. Low\footnotemark[2], Aviad Golan\thanks{MS Candidate, Stanford University, Department of Aeronautics and Astronautics, Stanford, CA 94305.}, Toby Bell\footnotemark[2], Antonio Rizza\thanks{Postdoctoral Fellow, Stanford University, Department of Aeronautics and Astronautics, Stanford, CA 94305.}, Simone D'Amico\thanks{Associate Professor, Stanford University, Department of Aeronautics and Astronautics, Stanford, CA 94305.}
}

\maketitle{}

\begin{abstract}
Spacecraft Rendezvous, Proximity Operations (RPO), and Formation Flying (FF) rely on safety-critical guidance, navigation and control (GNC) that must satisfy stringent performance and robustness requirements. 
However, verifying GNC performance is challenging due to the complexity and inaccessibility of the space environment, necessitating a verification and validation (V\&V) process that bridges simulation and real-world behavior. 
This paper contributes a unified, closed-loop, end-to-end digital and robotic twinning framework that enables software- and hardware-in-the-loop testing of spacecraft GNC systems. 
The framework is designed for modularity and flexibility, supporting interchangeable sensing modalities, control algorithms, and operational regimes.
The digital twin includes an event-driven faster-than-real-time simulation environment to support rapid prototyping.
The architecture is augmented with hardware-based robotic testbeds from Stanford's Space Rendezvous Laboratory (SLAB): the GNSS and Radiofrequency Autonomous Navigation Testbed for Distributed Space Systems (GRAND) to validate RF-based navigation techniques, and the Testbed for Rendezvous and Optical Navigation (TRON) and Optical Stimulator (OS) to validate vision-based methods.
The test article for this work is an integrated multi-modal GNC software stack developed at SLAB.
This paper introduces the hybrid twinning framework, summarizes calibration and error characterization of the robotic testbeds, and evaluates GNC performance across multiple operational modes in a full-range RPO scenario in LEO.
The results demonstrate consistency between software- and hardware-in-the-loop tests with clear explainability for deviations in performance, thus validating the hybrid twinning pipeline as a reliable framework for realistic assessment and verification of GNC systems.


\end{abstract}



\section{Introduction}

Autonomous Guidance, Navigation, and Control (GNC) is a key enabler for capabilities related to Rendezvous, Proximity Operations (RPO), Formation Flying (FF), and Distributed Space Systems (DSS) at large. 
Due to the inherent cost and inaccessibility of testing GNC software in space, it is best practice to validate these algorithms through the use of digital and robotic twins.
Digital twins are virtual models of the spacecraft and environment that are necessary for rapid prototyping numerical simulations. 
The process of using a digital twin to validate an algorithm is referred to as Software-in-the-Loop (SIL) testing. 
While SIL is a critical step in the GNC validation process, digital twins cannot fully capture the complexity and imperfections of flight hardware operating in the space environment.
To address this gap, Hardware-in-the-Loop (HIL) testing uses true or representative spacecraft hardware to both verify the digital twin's accuracy as well as evaluate the algorithm robustness under real-world conditions.


Although SIL and HIL testing are common practice in the space domain, existing frameworks are often limited in their scope and ability to test multiple GNC modalities across a full mission profile. 
In order to achieve the increasing level of autonomy and complexity required by future RPO and FF missions, it is necessary to develop more versatile, modular, and well-integrated approaches to GNC verification and validation (V\&V). 

To this end, this work implements a comprehensive hybrid software- and hardware-in-the-loop framework for closed-loop testing of spacecraft GNC software using digital and robotic twinning.
The digital twin of the spacecraft includes a high-fidelity dynamics  simulator, digital models of key sensors and  actuators, and an autonomous GNC flight software stack  used as a test article.
The high-fidelity simulator interfaces with vision- and radiofrequency (RF)-based sensor models to provide realistic sensor inputs, while the architecture also supports actuator models which interface with the simulator to update the spacecraft state.
The autonomous GNC software stack is multi-modal, modular, and can be replaced with another software stack as needed.

To enable HIL testing, hardware components are integrated into the architecture such that it operates as a hybrid digital and robotic twin of the spacecraft.
The hybrid architecture incorporates three complementary testbeds at Stanford's Space Rendezvous Laboratory (SLAB): the Testbed for Rendezvous and Optical Navigation (TRON)\cite{sharma_pose_2019, park2021tron} and Optical Stimulator (OS) \cite{beierle_design_2017, beierle2019high} for vision-based methods, the GNSS and Radiofrequency Autonomous Navigation Testbed for DSS (GRAND) \cite{giralo2018testbed} for RF-based navigation techniques.
Each hardware-based testbed may seamlessly replace components of the digital twin, enabling both rapid software prototyping and rigorous validation through hardware-based testing without significant additional overhead.

This novel approach supports end-to-end testing for multi-modal GNC stacks across complete mission scenarios. 
This work presents experimental results showcasing the multimodal capabilities of this testing framework by evaluating the GNC stack's performance across a representative mission scenario, including diverse navigation and control modes and validation using both digital and robotic twins of sensors. 



The remainder of the paper proceeds as follows. First, the ``\nameref{sec:state_of_the_art}" section summarizes the state-of-the-art in digital and robotic twinning for DSS. Then, the ``\nameref{sec:approach}" section presents the novel framework used in this work and describes the capabilities, calibration procedures, and resultant accuracy of the individual robotic twins. ``\nameref{sec:rpokit}" and ``\nameref{sec:experiments}" introduce the multi-modal GNC software stack and mission trajectory that are used for demonstrating the framework capabilities. Finally, the ``\nameref{sec:results}" section details the SIL and HIL results and how they complement each other to provide more robust validation. 


\section{Survey of Existing Digital and Robotic Twin Architectures}
\label{sec:state_of_the_art}

\subsection{Digital Twin State of the Art}

Space mission developers use a variety of software-based testing tools to simulate and model the space environment and verify flight algorithms before deployment. These include state-of-the-art general-purpose mission planning tools like Ansys's Systems Tool Kit (STK)\cite{ansys_stk}, GMAT \cite{hughes2016general}, and A.I. Solutions' FreeFlyer\textsuperscript{TM}\cite{freeflyer}, as well as flexible programmable modeling environments like Basilisk \cite{kenneally_basilisk_2020} and MATLAB/Simulink \cite{aerospace_toolbox}. Some groups also develop custom proprietary or internal simulation tools \cite{choo2004scibox, garcia2025dshell, penn2016trick } to meet their specific needs.

Mission planning tools (STK/GMAT/FreeFlyer) offer best-in-class high-fidelity models and trusted results, and are primarily designed for trajectory design and high-level mission analysis rather than end-to-end flight software validation. While they can include programmable extensions or scripting capabilities, they are often distributed as closed or semi-closed applications and impose rigid execution models, fixed data interfaces, and limited support for low-level integration with custom hardware and software.

Programmable environments (Basilisk/Simulink) offer more customization, allowing the user to easily specify new models and interfaces in Python or Matlab, and come closer to enabling integration with custom software and hardware. Basilisk in particular includes interfaces for comparing simulations with real-time hardware-in-the-loop results \cite{cols-margenet_modular_2016}. However, to maintain simplicity and convenience, these tools still provide limited control over timing, dynamic process scheduling \cite{basiliskscheduling}, and memory behavior, all of which are critical when validating embedded flight software against real hardware constraints.

With these limitations in mind, SLAB developed a novel event-driven simulation environment \cite{tbell2025sim}, described in the next section, as a backbone for the digital twin to address the timing and interfacing challenges of integrating with embedded flight software and disparate hardware testbeds. 
This environment was originally created to support GNC flight software validation for the VISORS mission \cite{guffanti2023autonomous} and was later extended to support the RPO stack developed by SLAB \cite{Kruger2024}.

\subsection{Hardware Testbed State of the Art}
Digital twins typically cannot fully reproduce realistic operating conditions due to modeling limitations.
Augmenting SIL tests with HIL ensures comprehensive end-to-end validation of the test article.
HIL testbeds serve as physical stand-ins for capabilities that are difficult to simulate accurately, helping validate both the algorithm and the digital twin itself. 
There are several types of HIL testbeds, including GNSS signal simulators, optical stimulators, kinematic spacecraft simulators, and dynamic spacecraft simulators \cite{wilde_historical_2019}. 

\subsubsection{GNSS Simulators:}

GNSS hardware testing captures fidelities in both the signal source and receiver end.
This includes the transmitter and receiver clock dynamics, tracking-loop transients, as well as packet latencies, which can only be modeled to limited fidelity in software-only emulation.
As such, GNSS HIL simulators serve a crucial role enabling high-fidelity and real-time validation of navigation flight software.
NASA Goddard Space Flight Center has developed a modular Formation Flying Test Bed (FFTB) since 2001 which eventually supported notable missions such as GRACE and MMS \cite{leitner2001gnsstestbed}. 
The FFTB was extended to integrate closed-loop end-to-end GNC \cite{burns2004gnsstestbed} in 2004. 
The German Space Operations Center (GSOC) and DLR too had developed their own native FFTB using a mission-independent Simulink-based architecture, to support rapid GNSS software prototyping \cite{gaias2012gnsstestbed}. 
Commercial options such as the Spirent GSS8000 \cite{fedora2008spirent} and IFEN NCS \cite{heinrichs2007ifen} are popular options for integration into existing testbeds.
Commercial products have supported development in multiple applications of spaceborne GNSS, such as (1) a precise orbit determination testbed for single spacecraft \cite{biswas2014gnsstestbed} at the University of New South Wales, (2) a testbed for ionospheric remote sensing using multi-constellation receivers in space \cite{peng2017gnsstestbed} at Virginia Tech, and (3) a precise formation flying testbed at SLAB \cite{giralo2018testbed}.
Presently, however, without bespoke solutions, commercial testbeds do not natively incorporate essential elements of a DSS mission simulation, such as live maneuver/actuator feedback as well as other software-emulated events such as crosslink, telemetry, and telecommanding feedback.

\subsubsection{Optical Stimulators:}
Optical stimulators are designed for component-level validation of vision-based sensors such as star trackers. 
These testbeds use collimated light sources to simulate distant objects, such as stars or spacecraft, on a digital screen. 
The rendered imagery is then used to stimulate the vision-based sensor under test. Examples of optical stimulators include SLAB's OS\cite{connor_os}, the John Hopkins University Applied Physics Laboratory's Celestial Object Simulator (CEL)\cite{CEL}, DLR's Jenoptik Optical Sky field Simulator (OSI) \cite{OSI}, Technical University of Denmark's Optical Stimulator System for VBS (OSVBS) \cite{roessler_optical_2014}, and the University of Naples' testbed \cite{rufino_real-time_2013}. 
While optical stimulators are compact, simple, and enable testing of GNC using real hardware, the spacecraft motion is still simulated in software only.
Additionally, they are limited in ability to account for narrower or wider field of views (FOV) when testing different vision-based sensors. 
Finally, they are often unable to simulate close-range scenarios when light from an object is no longer collimated \cite{beierle2019high}.

\subsubsection{Kinematic Testbeds:}
In contrast, kinematic testbeds physically model spacecraft motion by emulating commanded positions and orientations using high-torque actuators that are not representative of the true spacecraft actuators. 
They are primarily used for validating sensor performance and navigation algorithms, offering a modular and reconfigurable platform for testing vision-based GNC systems. 
Examples of existing kinematic testbeds include SLAB's TRON \cite{park2021tron}, DLR's European Proximity Operation Simulator (EPOS) \cite{boge2012epos, boge_new_2002} and EPOS 2.0 \cite{mietner_european_2017}, Lockheed Martin's Spacecraft Operations Simulation Center (SOSC) \cite{milenkovic2012lockheed}, DLR's Testbed for Robotic Optical Navigation (TRON) \cite{kruger_tron_2010}, and NASA Marshall Space Flight Center's Dynamic Overhead Target Simulator (DOTS) \cite{j_d_mitchell_automated_2007}. 
These testbeds include varying number of degrees-of-freedom (DOF), lighting conditions, and number of test articles.
However, kinematic testbeds are physically constrained by the size of the test facility, which often limits them to testing close-range GNC modalities.

\subsubsection{Dynamic Testbeds:}
Dynamic testbeds instead model the true spacecraft dynamics by using realistic actuation and a facility designed to minimize interaction between the test article and the local environment.
A common approach is to use air-bearing tables, which simulate 3-DOF motion (2 translational and 1 rotational DOF) by floating spacecraft models on a thin film of pressurized gas.
Examples include the Naval Postgraduate School's Proximity Operations of Spacecraft: Experimental hardware-In-the-loop DYNamic simulator (POSEIDYN) \cite{POSEIDYN}, NASA JPL's Small Satellite Dynamics Testbed (SSDT) \cite{sternberg2018jet, jpl_newer},  Naval Postgraduate School's Three-Axis Simulator (TAS-2) \cite{tappe_development_2009}, and Stanford University's Free Flyer Testbed \cite{mote_collision-inclusive_2020}. 
Additional facilities combine multiple platforms using flat air-bearing tables and spherical-bearing tables to achieve 5 or 6 DOF, such as Georgia Institute of Technology's Autonomous Spacecraft Testing of Robotic Operations in Space (ASTROS) testbed \cite{GTechAstros}.
Finally, other testbeds use a combination of kinematic and dynamic approaches to achieve full 6-DOF motion, such as California Institute of Technology's Multi-Spacecraft Testbed for Autonomy Research (M-STAR) \cite{nakka_six_2018}.
While dynamic testbeds capture certain dynamics capabilities such as contact dynamics, they are less modular and more operationally complex, requiring fully self-contained test articles with onboard power, actuation, and computation.
Like kinematic testbeds, dynamic testbeds are limited by facility size and thus cannot be used for testing far-range GNC capabilities.

Each type of HIL testbed offers key validation capabilities over pure SIL testing and serves a distinct role in the GNC validation pipeline. However, they also introduce practical constraints such as the type of GNC modes they can support (e.g. vision-based versus RF-based navigation), the inter-satellite distances they can replicate, and the fidelity of spacecraft dynamics they can achieve. Thus, no testbed alone can support multi-modal GNC V\&V across a full-range trajectory.

\section{Hybrid Digital and Robotic Twinning Approach}
\label{sec:approach}

To overcome the limitations of existing testbeds and tools, this work introduces a hybrid digital and robotic twinning framework that enables systematic validation of spacecraft GNC systems. The presented approach supports progressive V\&V, from rapid SIL testing to fully integrated HIL experiments. 

The digital twin captures the core components of the spacecraft system and environment, including a high-fidelity dynamics simulator, sensor and actuator models, and the full GNC software stack. 
Its sensor models consist of an OpenGL-based renderer coupled with a camera model and a target spacecraft CAD model to simulate optical measurements and a GNSS Software Emulator for GNSS measurements.
In total, this digital twin enables testing of far-, \mbox{mid-,} and close-range vision-based and radio-frequency (RF) based navigation algorithms in closed-loop with control. 
Because all components are software-based, the digital twin enables rapid and repeatable simulations for both algorithm development and initial validation.
 
The robotic components extend the capabilities of the proposed framework by replacing one or more of the simulated sensor models with physical hardware, enabling the architecture to operate as a hybrid of digital and robotic models. 
Specifically, the robotic components consist of three complementary testbeds: the Optical Stimulator (OS) \cite{beierle_design_2017,beierle2019high}, the Testbed for Rendezvous and Optical Navigation (TRON) \cite{sharma_pose_2019, park2021tron}, and the GNSS and Radiofrequency Autonomous Navigation Testbed for Distributed Space Systems (GRAND) \cite{giralo2018testbed}. 
Each testbed is tailored for different sensing modalities and inter-satellite distances (ISD), where ISD refers to the relative separation of the target and chaser spacecraft. For vision-based navigation algorithms, the ISD can be categorized as far-, mid-, or close-range based on the sensor's ability to resolve the target in the image. 
In far-range imagery, the target is unresolved and appears as a point source. 
Mid-range refers to semi-resolved imagery where the target is no longer a point source but without any clearly distinguishable features. 
Finally, at close-range separations, the imagery is fully resolved, allowing spacecraft features to be identified and leveraged for navigation. 

The OS is used to test far-, mid-, and close-range vision-based navigation algorithms using star trackers and other monocular cameras coupled with synthetically rendered space scenes. 
TRON is used for testing close-range navigation algorithms using an optical payload, a scaled model of the target, and realistic illumination conditions to create pseudo-spaceborne imagery.
In contrast, RF-based navigation does not require target imagery and therefore is applicable across all ISD regimes. GRAND enables testing of RF-based navigation by combining a multi-GNSS signal generator coupled with numerical simulation of ground truth dynamics.
The integration of these three testbeds enables coverage of multi-modal GNC HIL validation across a complete range of inter-satellite distances. Table~\ref{tab:twins_overview} summarizes the robotic testbed capabilities and the corresponding digital twin components they replace. 

\begin{table}[]
    \centering
    \caption{High-level overview of the robotic testbeds used for hardware-in-the-loop validation within the hybrid digital and robotic twinning framework.}
    \small
    \begin{tabular}{lcccc}
    \toprule
    \makecell{\textbf{Robotic} \\ \textbf{Testbed}} &
    \textbf{ISD Range} &
    \makecell{\textbf{Sensing} \\ \textbf{Mode}} &
    \makecell{\textbf{Hardware} \\ \textbf{Components}} &
    \makecell{\textbf{Digital Twin} \\ \textbf{Sensor Analog}} \\
    \toprule
    \textbf{OS} & \makecell{Far \\ Mid \\ Close} & \makecell{Vision} & Optical Payload & \makecell{OpenGL-Based Renderer} \\
    \hline
    \textbf{TRON} & \makecell{Close} & \makecell{Vision} & \makecell{Optical Payload \\ Spacecraft Model \\ Albedo Light boxes} & OpenGL-Based Renderer \\
    \hline 
    \textbf{GRAND} & \makecell{Far \\ Mid \\ Close} & RF & \makecell{GNSS Signal Generator \\ GNSS Receiver} & GNSS Software Emulator \\
    \bottomrule
    \end{tabular}
    \vspace{-4mm}
    \label{tab:twins_overview}
\end{table}

\begin{figure}[htb]
    \centering
    \includegraphics[width=\linewidth]{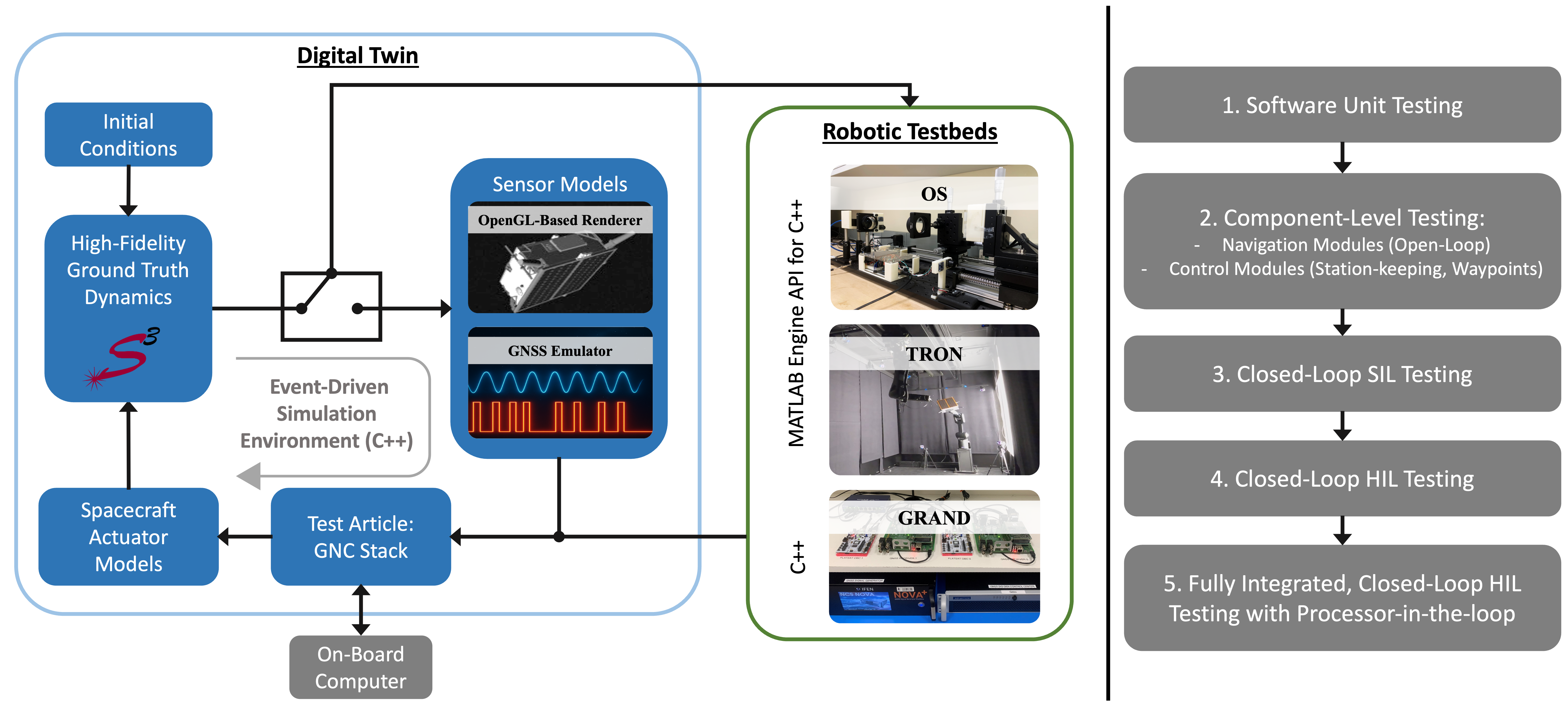}
    \caption{(\emph{Left}) Hybrid digital and robotic twinning framework. Details of the operating interfaces for the OS, TRON, and GRAND can be found in Figures~\ref{fig:OS_testbed}, \ref{fig:TRON}, \ref{fig:ifen-trash}, respectively.  (\emph{Right}) The incremental GNC V\&V testing methodology used in this work.}
    \label{fig:HIL_SIL_framework}
\end{figure}

An overview of the hybrid framework is shown on the left side of Figure~\ref{fig:HIL_SIL_framework}.
All SIL and HIL tests are driven by a common source of ground truth dynamics, initialized by the user-defined initial conditions. 
The ground truth dynamics are fed into either the digital twin's sensor models or the corresponding robotic testbed, which generates measurements that are used as inputs for GNC.
The GNC software may be executed on the same host computer as the simulation to isolate the algorithm performance, or can alternatively be embedded on an on-board computer (OBC) to additionally account for the compute constraints of flight hardware. 
The GNC stack responds with control inputs as necessary, which are carried out by the digital twin's actuator models, thus updating the ground truth state. 
A key feature of this framework is the use of composable sensor interfaces that enable seamless substitution between the digital sensor models and robotic testbeds. 
From the perspective of the GNC stack, the sensor measurements are returned through a common interface, independent of whether they originate from software or hardware. 
Switching between SIL and HIL operation therefore only requires setting a configuration flag in the top level simulation. 
When a robotic component is selected, the measurements are routed through the testbed's operating software with minimal user interaction.

The hybrid digital and robotic twinning architecture enables a structured and incremental GNC V\&V workflow, illustrated on the right side of Figure~\ref{fig:HIL_SIL_framework}. Testing progresses from low-level software unit tests to fully integrated systems-level assessments with processor-in-the-loop execution. 
The modular structure allows individual GNC modules, such as specific navigation algorithm or controllers, to be tested in isolation in a consistent environment before being integrated into a closed-loop simulation. 
Moreover, the framework supports controlled introduction of real hardware, enabling systematic assessment of algorithm sensitivity to sensor imperfections and environmental uncertainties.
As a result, the proposed approach is well suited for both early-stage algorithm development and later-stage system-level validation. 

\subsection{Digital Twin}
\label{sec:digital_twin}

The backbone of the digital twin is a C++ event-driven simulation environment, which drives the ground truth dynamics and regulates the flow of data by queuing and executing simulation events, maneuvers, measurements, telemetry and telecommands with realistic latencies \cite{tbell2025sim}.
The ground truth orbit dynamics are propagated using SLAB's high-fidelity astrodynamics library, $\mathcal{S}^3$ \cite{giralo2018testbed}.
The simulation integrates Gauss Variational Equations (GVE) using the Dormand–Prince method\cite{dormand_family_1980} with an adaptive step size to bound truncation error. 
The modeled perturbations include a 60x60 GGM05S spherical harmonic gravity model, the NRLMSISE-00 atmospheric model \cite{picone_nrlmsise-00_2002} with a cannonball drag model, and solar radiation pressure (SRP) with a cylindrical Earth shadow. 
Currently, the ground truth propagation additionally includes perturbation-free attitude dynamics to simulate a tumbling target spacecraft, which is representative of short-duration uncontrolled motion. 

The digital twin includes a library of measurements and actuation models designed for rapid, iterative development of GNC stacks for distributed space systems.
Imaging sensors are simulated using a high-fidelity OpenGL-based renderer.
The synthetic images are generated in OpenGL using SLAB's Optical Stimulator software \cite{beierle2019high} given a spacecraft CAD model and camera model.
The renderer can simulate non-resolved point sources as well as fully resolved objects. 
It includes an accurate star field and configurable illumination conditions with both specular and diffuse reflections.

The GNSS Software Emulator provides a faster-than-realtime alternative to the HIL GRAND testbed by augmenting the event-driven simulation environment \cite{tbell2025sim} with a GPS constellation of 31 space vehicles.
It emulates pseudorange and carrier phase measurements packaged into NovAtel-format binaries \cite{novatelmanual}, corrupted with ionospheric effects on the group delay and phase advance, carrier phase ambiguities, clock errors, thermal noise and body frame offsets due to the Center-of-Mass (COM) to Phase Center Offset (PCO).
All of the parameters for the GNSS Emulator are included in \autoref{tab:gnss-emu-sig} and \autoref{tab:gnss-emu-sc} in \nameref{sec:appendix_1}.

The SIL emulated crosslink, which enables mutual exchange of GNSS measurements between cooperative spacecraft, models stochastic transmission delay with message delivery failure.
This can expose weaknesses resulting from non-robust navigation algorithms with a reliance on shared measurements, common in Carrier Phase Differential GNSS (CDGNSS).
This environment supports deterministic, faster-than-real-time execution, enabling developers to simulate multi-spacecraft missions, debug complex interactions, and identify implementation-level defects, e.g., memory exhaustion or communication faults which are capabilities not found in traditional HIL setups \cite{tbell2025sim}. 

Finally, the digital twin incorporates the GNC flight software stack under evaluation.
This component will be detailed  in the section ``\nameref{sec:rpokit}''.
 
\subsection{Optical Stimulator (OS)}
\label{sec:os}
\begin{figure}[t!]
    \centering
    \includegraphics[width=1.0\linewidth]{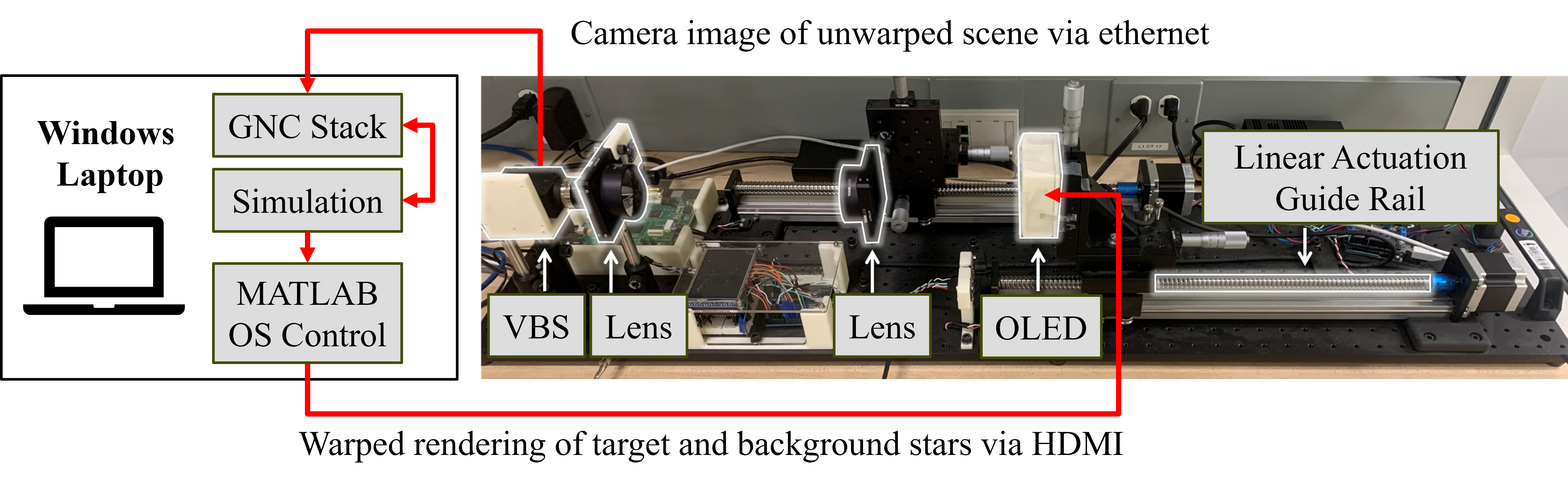}
    \caption{Closed loop operation of the Optical Stimulator testbed. Figure recreated from Ref.~\citenum{connor_os}.}
    \label{fig:OS_testbed}
\end{figure}

The OS is a variable-magnification optical bench designed to stimulate a wide range of vision-based sensors using synthetically generated space scenes, replacing the camera model used in the digital twin with real hardware to capture sensor noise and distortion that are difficult to model accurately.
The testbed consists of a vision-based sensor (VBS) test article, two corrective lenses, and an organic light-emitting diode (OLED) microdisplay as shown in Figure \ref{fig:OS_testbed}. 
The lenses and OLED monitor can be moved along a rail using motorized lead screws to vary the magnification of the scene depending on the VBS FOV.

In this work, the VBS is a PointGrey GrassHopper3-PGE-23S6M-C with a Xenoplan 1.9/35 lens, a FOV of $18.3 ^\circ$, and a resolution of 1920$\times$1200 pixels.
However, the platform allows users to use any camera suited to the relevant mission, such as a star tracker.
The OS renders a space scene on the OLED monitor, the scene is magnified through the lenses to fill the camera's FOV, and the image is captured by the mounted VBS. 
The OLED display and camera are operated by a MATLAB program running on a host computer.
During HIL testing, the C++ simulation calls the MATLAB OS control program through the MATLAB Engine API at the prescribed imaging rate, triggering the camera to take an image which is returned via Ethernet.
The simulator supports various imaging rates by pausing the simulation until an image is returned to the GNC stack.

Ref.~\citenum{connor_os} details the OS calibration process, which is designed to ensure geometric and radiometric accuracy.
The geometric calibration process removes the corrective lens distortion and tunes how the synthetic scene is rendered on the OLED monitor.
The radiometric calibration characterizes the irradiance contribution from a single OLED pixel.
Together, these steps ensure that light will reach the VBS aperture from the desired angular location with the correct visual magnitude.
To validate the distortion calibration parameters, a set of known unit vectors are used to display a grid of dots on the OLED monitor after their directions are warped using the optimal distortion coefficients introduced by the corrective lenses. When the VBS captures an image, the measured grid locations should result in the known, given unit vectors again. Calibration with the PointGrey produces residuals with mean 10.17 arcseconds, which is on par with calibration accuracy reported in Ref.~\citenum{beierle2019high}. 


\subsection{Testbed for Rendezvous and Optical Navigation (TRON)}
\label{sec:tron}
TRON is a robotic testbed for verification of vision-based navigation at close range.
TRON features two KUKA industrial robots which simulate kinematic motion of a chaser spacecraft and a target object. 
The robots can achieve precise positions with 0.03 mm motion repeatability.
The first robot is a 7-degree-of-freedom (7-DOF) robotic manipulator that travels along a 7 m ceiling-mounted track, allowing for up to 6 meters of separation between the target and the camera, though the simulated distance can be increased by using a scale model. 
A space-capable Point Grey Grasshopper 3 camera with a Xenoplan 1.4/17 mm lens is mounted on the rail-mounted robot's end-effector. 
Like the OS, this camera can be replaced with another VBS as needed.
The second robot is a 6-DOF grounded robotic arm, equipped with a mount which allows the attachment of a mockup model of the target. 
Each robot reports internal joint angles; through forward kinematics, the position and orientation of its end effector is known with respect to a origin defined in the KUKA reference system. 
The TRON facility also includes a sun lamp and 10 wall-mounted Earth albedo lightboxes which are controlled together to accurately simulate the illumination conditions of space.
All surfaces are painted with a radiation-absorbing paint to mitigate unwanted reflections in the visible range. 

\begin{figure}[ht]
    \centering
    \includegraphics[width=0.9\linewidth]{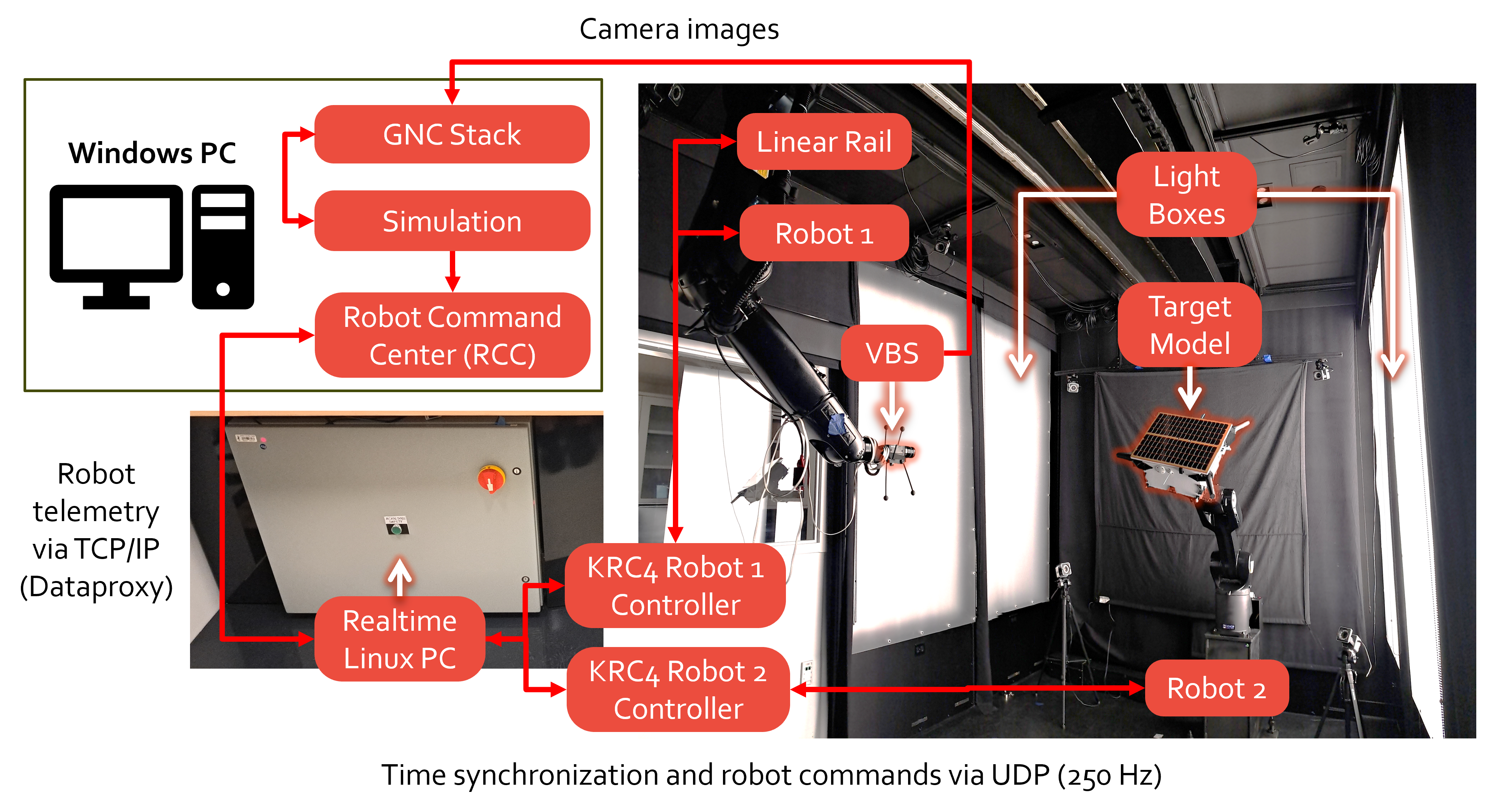}
    \caption{TRON Facility. Figure recreated from Ref.~\citenum{park2021tron}.}
    \label{fig:TRON}
\end{figure}

The TRON facility also features 12 Vicon IR motion capture cameras mounted on ceiling rafters and tripods to provide full coverage of the facility. 
The Vicon cameras track reflective spheres attached to the target and camera and report their positions and orientations with respect to a local reference frame. 
The  Vicon system provides secondary measurements that are independent from the robots, increasing TRON's calibration accuracy.

Like the OS, TRON is operated through a MATLAB function invoked by the simulation.
The function receives a relative pose and lighting direction as inputs, and returns an image from the VBS for use by the GNC stack. 
Specifically, it uses the inputs to compute the desired robot positions in joint-angle space along with lightbox settings.
The resulting joint angles are converted into a motion trajectory satisfying dynamic constraints and transmitted to the robot controllers via KUKA's Robot Sensing Interface (RSI) over a 250 Hz UDP connection. 
The KUKA robot end-effector achieved positions are then reported back to the host computer via TCP/IP. 
Once the desired position is achieved, the function triggers the VBS to take an image, which is returned to the host computer via Ethernet.
When the image is passed back to the GNC stack, a helper function applies a boolean mask to replace background pixels with a noisy black background.

TRON's calibration process begins with an absolute calibration of the robots, where the robots' joint offsets are estimated for precise positioning. 
A camera attached to the ceiling mounted robot collects images of a ChArUco board\cite{garrido2014automatic} attached to the other robot.
Images are taken to maximize the diversity of relative poses, even if these poses would not be experienced during regular operations. 
The relative pose information from the robots, Vicon, and via solving the Perspective-n-Point (PnP) for the ChArUco board is used to estimate the robot joint offsets via nonlinear least-squares. 
A correction to the joint offsets is then calculated and applied to the robot commands.
The second calibration step for TRON involves estimating the relative position and orientation between the robot end effectors and the mounted objects, i.e. the camera or target, by solving the Robot-World-Hand-Eye (RWHE) calibration problem for each measurement source \cite{tabb_solving_rwhe}. Ref \citenum{park2021tron} provides details on this calibration process, which uses the same measurement sources as before and performs Bayesian data fusion to increase the accuracy of the estimated offsets.

The calibration results of this work are shown in Table \ref{tbl:rwhe_results}. 
Here, $E_T$ is the relative translation error, $E_R$ is the relative rotation error, and $E_p$ is the pixel error of the ChArUco board's markers. 
Because the TRON facility currently uses a half-scale target model, $E_T$ as computed during calibration is half of the ``effective" $E_T$ corresponding to a full-scale target.
The results reported here have slightly lower accuracy compared to previously published results with this same method and target model\cite{d2012prisma}. 
This is due to a larger coverage of relative poses in this work, while previous calibrations were done in a limited region of the facility. 

\begin{table}
\centering
\small
\caption{Robot World Hand-Eye (RWHE) Calibration Results (Mean ± Std)}
\begin{tabular}{cccc}
\toprule
Metrics & KUKA-only RWHE & Vicon-only RWHE & Data Fusion \\ 
\midrule
$E_\textrm{T}$ [mm] & 3.968 $\pm$ 4.295 & 6.661 $\pm$ 8.273 & 2.894 $\pm$ 3.132 \\ 
$E_\textrm{R}$ [${}^\circ$] & 0.369 $\pm$ 0.584 & 0.364 $\pm$ 0.557 & 0.350 $\pm$ 0.566 \\ 
$E_\textrm{p}$ [pix] & 3.418 $\pm$ 1.775 & 9.968 $\pm$ 6.610 & 3.244 $\pm$ 1.796 \\
\bottomrule
\label{tbl:rwhe_results}
\end{tabular}
\vspace{-4mm}
\end{table}

\subsection{GNSS and Radiofrequency Autonomous Navigation Testbed for DSS (GRAND)}
\label{sec:grand}

\begin{figure}[h]
    \centering
    \includegraphics[width=1.0\linewidth]{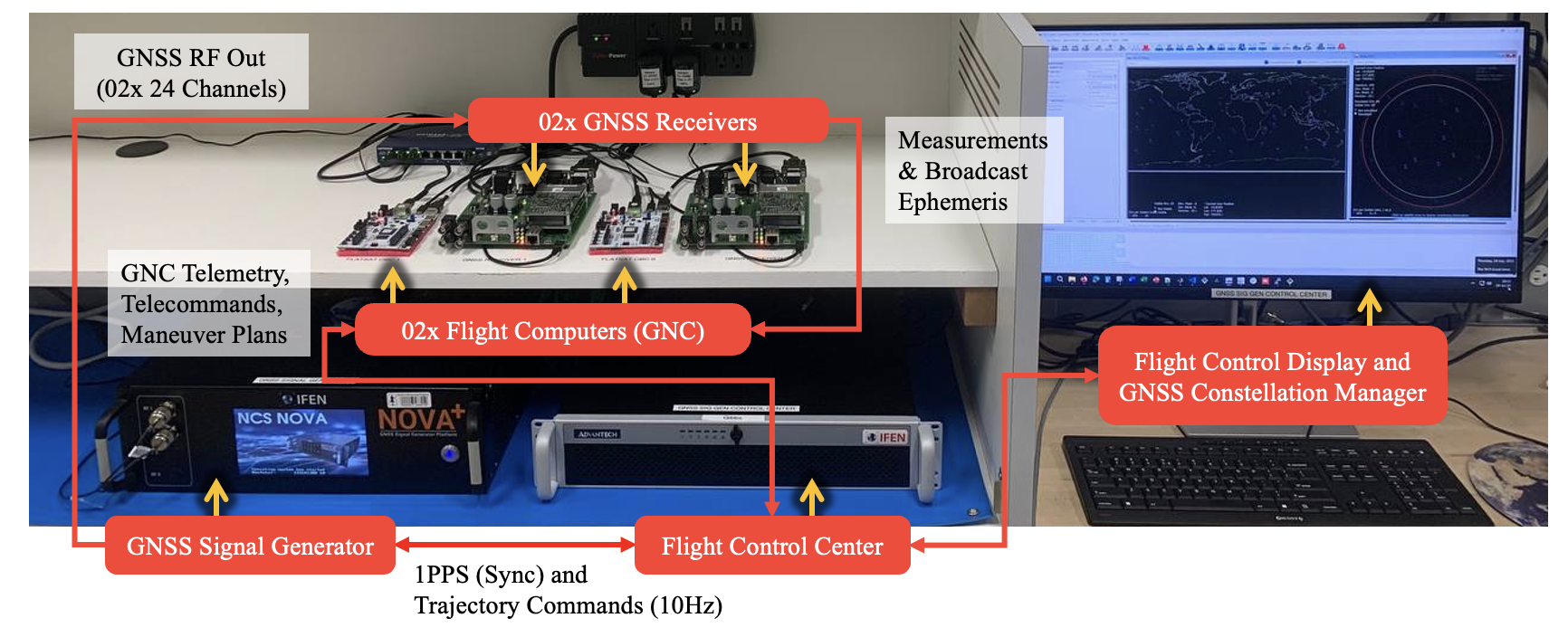}
    \caption{The GRAND testbed, comprising (i) an IFEN NOVA+ 2020 multi-GNSS signal generator, (ii) the ground flight control center, and (iii) a set of NovAtel OEM628 GNSS receivers, with flight heritage, connected serially to ZYNQ7000 SoC boards equivalent to the VISORS\cite{guffanti2023autonomous} flight computers. Both GNSS receiver and flight computer models possess flight heritage. The ZYNQ7000 flight computers can be used to host C-compatible GNC software test articles.}
    \label{fig:ifen-trash}
\end{figure}

The GRAND testbed of \autoref{fig:ifen-trash} enables real-time evaluation of GNSS-based flight software and navigation algorithms.
GRAND was originally conceived to test the Distributed multi-GNSS Timing and Localization (DiGiTaL) software package to achieve precise relative navigation between cooperative spacecraft using CDGNSS with Integer Ambiguity Resolution (IAR) \cite{teunissen1994lambda}.
It has since been upgraded from its original design \cite{giralo2018testbed} in three ways.
First, the original IFEN NavX has been upgraded to an IFEN NOVA+, with greater channel bank support and hardware compatibility with modern multi-GNSS signals.
Second, the flight control center has a more powerful Intel Core i7-14700K (3.40 GHz) processor to support simulating more sophisticated ground-based tasks.
Third, the testbed is fully integrated with the event-driven simulation loop described in ``~\nameref{sec:digital_twin}"~\cite{tbell2025sim}, which operates in tandem with the flight control center and the constellation manager.
This integration enforces a single notion of time across live-streamed hardware-generated GNSS messages and all software-simulated events.
This includes orbital and attitude propagation, emulated inter-satellite crosslinks, maneuver planning and execution, actuator dynamics, propulsion telemetry, and scheduled communication latencies.
This time synchronization allows the testbed to preserve causal consistency when running the GNC stack in closed-loop feedback with the ground truth measurement and dynamics sources.

The flight control center is synchronized with the IFEN NOVA+ via a 1PPS line.
Ground truth dynamics are fed to the IFEN NOVA+ over a UDP socket at \SI{10}{\hertz}, which provides a flight trajectory that is interpolated for fine-sampling.
The IFEN NOVA+ produces RF GNSS signals from a simulated GNSS constellation, so that the receiver measurements are consistent with the input flight trajectory.
The GNSS receivers in turn provide measurements and ephemeris to the payload GNC software, which can optionally be hosted on the host computer or representative flight computers.
GRAND has successfully validated real-time execution of the VISORS GNC stack \cite{guffanti2023autonomous} and continues to serve as a critical testbed for CDGNSS with IAR \cite{low2024coupling, low2026digital}.

\section{Autonomous GNC Stack for RPO}
\label{sec:rpokit}
The test article for this work is an autonomous GNC stack for RPO and FF that supports operations across all mission phases, from initial acquisition to final approach, using a unified framework deployed on one or more ``chaser" spacecraft. The full details for this GNC stack are described in Ref.~\citenum{Kruger2024}, while the high-level architecture and algorithms are summarized here.

The GNC architecture comprises four core modules: management, navigation, safety, and control. The management module handles operational mode transitions and fault detection, isolation, and recovery (FDIR). 
The navigation module fuses both classical and machine learning-based measurements from optical sensors, GNSS, and inter-satellite links to estimate relative and absolute states of targets using both batch initialization and sequential estimation.
The safety module predicts potential collisions based on state estimates and provides risk metrics to the control system. 
The control module generates maneuver plans through a suite of optimization-boosted solver algorithms, enabling autonomous trajectory planning, station keeping, and collision avoidance. 
A subset of navigation and control algorithms from this GNC stack are employed in the experiments that will be detailed in the next section and are summarized in Tables~\ref{tab:rpokit_nav} and \ref{tab:rpokit_ctrl}. 

The architecture is designed to execute fully onboard, allowing servicers to autonomously navigate, manage system and safety constraints, and plan control actions in real time. 
Collision risk assessments, expressed in the servicer's radial-tangential-normal (RTN) frame, inform maneuver planning to ensure safe operation even under degraded conditions.
Informed by the safety assessment and the estimated state, the control module uses an internal state machine to determine whether it is in a waypoint approach or station-keeping mode, then employs an appropriate control solver to achieve the control objective.

\begin{table}[t]
\scriptsize
\centering
\caption{Navigation and estimation algorithms tested in this work. Algorithms are a subset of the Autonomous GNC stack from Ref.~\citenum{Kruger2024}. Table recreated from Ref.~\citenum{Kruger2024}.}
\begin{tabular}{l|llll}
\toprule
\textbf{Algorithm} & \makecell[l]{\textbf{Angles-Only} \\ \textbf{Tracking (AOT)}\cite{kruger2021autonomous}} & \makecell[l]{\textbf{Spacecraft Pose} \\ \textbf{Network (SPN)}\cite{park_robust_2023}} & \makecell[l]{\textbf{CDGNSS} \\ \textbf{with IAR}\cite{giralo2019digital, low2026digital}} & \makecell[l]{\textbf{Sequential Estimation} \\ \textbf{and Data Fusion (SEDF)}\cite{sullivan_generalized_2021, park_adaptive_2023}} \\
\hline
\makecell[l]{\textbf{Sensing} \\ \textbf{Modality}}
& Vision-based 
& Vision-based
& RF-based
& - \\
\hline
\makecell[l]{\textbf{Relevant} \\ \textbf{Techniques}} 
& \makecell[l]{Multi-hypothesis \\ tracking \\ Kinematic target \\ modeling} 
& \makecell[l]{Convolutional \\ neural networks} 
& \makecell[l]{mLAMBDA \\ algorithm \cite{teunissen1994lambda, chang2005mlambda}}
& \makecell[l]{Unscented Kalman filter\\ Statistical data association} \\
\hline
\makecell[l]{\textbf{Possible} \\ \textbf{Inputs}} 
& \makecell[l]{Host camera images\\ Host orbit estimate} 
& \makecell[l]{Host camera images\\ Host orbit estimate\\ Host attitude estimate} 
& \makecell[l]{Host GNSS\\ Peer GNSS} 
& \makecell[l]{Host \& peer bearing angles\\ Host \& peer GNSS\\ Host \& peer camera attitude\\ Host SPN heatmaps\\ Host \& peer state estimates\\ Host \& peer maneuver plan} \\
\hline
\makecell[l]{\textbf{Possible} \\ \textbf{Outputs}} 
& \makecell[l]{Target bearing angles\\ Sensor attitude} 
& \makecell[l]{Estimated target \\ keypoint locations \\ } 
& \makecell[l]{Resolved double- \\ differenced \\ integer ambiguities} 
& \makecell[l]{Host absolute orbit estimate\\ Peer relative orbit estimates\\ Host absolute attitude estimate\\ Peer relative attitude estimates\\ Host \& peer auxiliary state est.} \\
\hline
\makecell[l]{\textbf{Nominal Sample} \\ \textbf{Time (LEO)}} 
& 60 sec & 5 sec & 10 sec & 5–120 sec \\
\hline
\makecell[l]{\textbf{Representative} \\ \textbf{Accuracy (1-$\sigma$)}} 
& \makecell[l]{$<$35" angle error\\ $<$20" attitude error} 
& \makecell[l]{$<$20 cm relative pos.\\ $<$10$^\circ$ orientation} 
& \makecell[l]{$<$1 cm relative \\ position error}
& \makecell[l]{$<$5 m abs. pos. (GNSS w/out IAR)\\
  $<$5 cm rel. pos. (GNSS w/out IAR)\\
  $<$1 cm rel. pos. (GNSS with IAR)\\
  $<$1 km abs. pos. (angles-only)\\
  $<$2\% $\delta\lambda$ rel. pos. (angles-only)\\
  $<$10 cm rel. pos. (SPN)\\
  $<$1° rel. att. (SPN)} \\
\hline
\textbf{Applicability} 
& Non-resolved targets 
& \makecell[l]{Resolved targets \\ w/ pre-selected \\ keypoints} 
& Cooperative targets 
& - \\
\bottomrule
\end{tabular}
\vspace{-4mm}
\label{tab:rpokit_nav}
\end{table}


\begin{table}[ht]
\small
\centering
\caption{Controls algorithms tested in this work. Algorithms are a subset of the Autonomous GNC stack from Ref.~\citenum{Kruger2024}. Table recreated from Ref.~\citenum{Kruger2024}.}
\begin{tabular}{c|lll}
\toprule
\textbf{Control Solver} & \makecell[l]{\textbf{Chernick-D'Amico} \\ \textbf{Closed-Form}\cite{chernick_closed-form_2021, chernick_new_2018}} & \makecell[l]{\textbf{Koenig-D'Amico} \\ \textbf{SOCP/QP}\cite{koenig2020fast}} & \makecell[l]{\textbf{Least Squares}\cite{guffanti2023autonomous}} \\
\hline
\makecell[c]{\textbf{Solution Type}} 
& \makecell[l]{Impulsive} 
& \makecell[l]{Optimized impulsive} 
& \makecell[l]{Time-fixed impulsive} \\
\hline
\textbf{Applicability} 
& \makecell[l]{Large ISD \\ Station-keeping \\ Reconfiguration}
& \makecell[l]{Small ISD \\ Trajectory Tracking} 
& \makecell[l]{Small ISD \\ Trajectory Tracking} \\
\hline
\makecell[c]{\textbf{Relevant} \\ \textbf{Techniques}} 
& \makecell[l]{Reachable Set Theory} 
& \makecell[l]{Reachable Set Theory \\ Convex optimization} 
& \makecell[l]{Least Squares with \\ Tikhonov regularization} \\
\hline
\makecell[c]{\textbf{Maneuver} \\ \textbf{Type}} 
& \makecell[l]{3 T, 1-2 N} 
& \makecell[l]{4-6 RTN} 
& \makecell[l]{2-3 RTN} \\
\hline
\makecell[c]{\textbf{Maneuver} \\ \textbf{Frequency}} 
& \makecell[l]{0.5 Orbits} 
& \makecell[l]{Optimized} 
& \makecell[l]{Time-Fixed} \\
\hline
\makecell[c]{\textbf{Planning} \\ \textbf{Horizon}} 
& $>$3 orbits & 0.1 $<$ orbits $<$ 3 & $<$3 orbits \\
\hline
\makecell[c]{\textbf{Control Loop} \\ \textbf{Closure Frequency}} 
& 300 sec & 60 sec & 60 sec \\
\hline
\makecell[c]{\textbf{Nominal} \\ \textbf{Accuracy}} 
& 1-10 m & 1-10 cm & 1-10 cm \\
\bottomrule
\end{tabular}
\vspace{2mm}
\label{tab:rpokit_ctrl}
\end{table}

\section{Experiments}
\label{sec:experiments}

After completing software unit testing and component-level tests as shown in Figure \ref{fig:HIL_SIL_framework}, three experiments are conducted in closed-loop SIL and HIL to demonstrate the testing framework's ability to evaluate the GNC stack's performance over an entire RPO mission and across a broad range of testing and GNC modalities.
To this end, this work uses an RPO trajectory spanning both far and close separation distances.
The RPO trajectory consists of two spacecraft: a servicer spacecraft and a target.
The target used in this work is the Tango spacecraft from the PRISMA mission. \cite{damico_noncooperative_2013}.

\subsection{RPO Trajectory}

The servicer's absolute orbit is a dawn-dusk Sun-synchronous low-Earth orbit (LEO) with an 18h nominal Local Time at the Ascending Node (LTAN) derived from the PRISMA mission \cite{d2012prisma} \cite{park_adaptive_2023}. 
Such an orbit is ideal for servicing missions due to consistent lighting conditions. 
The orbit is initialized with the quasi-nonsingular orbital elements listed in Table \ref{tab:initial servicer elements} and an initial epoch of July 18, 2011 at 01:00:00 UTC.

\begin{table}[h]
\centering
\caption{Servicer Quasi-nonsingular Orbital Elements}
\begin{tabular}{c c c c c c}
    \hline
    $a$ [km] & $e_x$ [-] & $e_y$ [-] & $i [^\circ]$ & $\Omega [^\circ]$ & $u [^\circ]$ \\
    \hline
    7,087.30 & 0.0015 & $-4.17 \times 10^{-8}$ & 98.18 & -90 & -0.69
\label{tab:initial servicer elements}
\end{tabular}
\end{table}

The reference trajectory consists of waypoints that parameterize the target's relative orbit with respect to the servicer at certain points in time, given in the six quasi-nonsingular relative orbital elements (ROE) as defined by D'Amico \cite{damico}. 
The complete set of mission waypoints are listed in Table \ref{tab:full range waypoints}, including an approach from far range (waypoints \#0 to \#7) to a close approach near the target for inspection (waypoints \#8 to \#10).

\begin{table}[h!]
\centering
\caption{End-to-end Mission ROE Waypoints}
\begin{tabular}{ccrrrrrr}
\textbf{Waypoint} & \textbf{Time (hr)} & $a\delta a$ (m) & $a\delta \lambda$ (m) & $a\delta e_x$ (m) & $a\delta e_y$ (m) & $a\delta i_x$ (m) & $a\delta i_y$ (m) \\
\hline
0  & 0 & -2000.0 & -75000.0 & 0.0 & 0.0 & 0.0 & 0.0 \\
1  & 6.0 & -87.0  & -50000.0  & 0.0   & 1000.0 & 0.0   & 1000.0 \\
2  & 12.0  & -70.0  & -25000.0  & 0.0   & 500.0  & 0.0   & 500.0  \\
3  & 18.0  & -17.0  & -5000.0   & 0.0   & 500.0  & 0.0   & 500.0  \\
4  & 24.0  & 0.0  & -1000.0   & 0.0   & 30.0  & 0.0   & 30.0  \\
5  & 28.0  & 0.0  & -600.0   & 0.0   & 18.0  & 0.0   & 18.0  \\
6  & 32.0  & 0.0  & -200.0   & 0.0   & 6.0  & 0.0   & 6.0  \\
7  & 36.0  & 0.0  & -100.0   & 0.0   & 3.0  & 0.0   & 3.0  \\
8 & 39.0 & -0.5 & -10.0 & 0.0 & 2.0 & 0.0 & 1.0 \\ 
9 & 40.0 & 0.0 & -7.0 & 1.0 & 0.0 & 2.0 & 0.0 \\
10 & 41.25 & 0.0 & -8.5 & 1.0 & 0.0 & 1.0 & 0.0 \\ 

\end{tabular}
\vspace{-5mm}
\label{tab:full range waypoints}
\end{table}

The target spacecraft begins at far range, approximately 75 km behind the servicer in the along-track direction while approaching at a rate of 3.2 m/s (waypoint \#0). 
Then, the rate of approach is reduced, and relative e-/i- vector separation \cite{d2006proximity, damico} is added to ensure passive safety (waypoints \#1 to \#3). 
The intersatellite distance is incrementally decreased from 1 km to 100 m (waypoints \#4 to \#7). 
At these waypoints, the target is unresolved or semi-resolved when observed by the VBS used in this work. 

\begin{figure}
    \centering
    \includegraphics[width=0.9\linewidth]{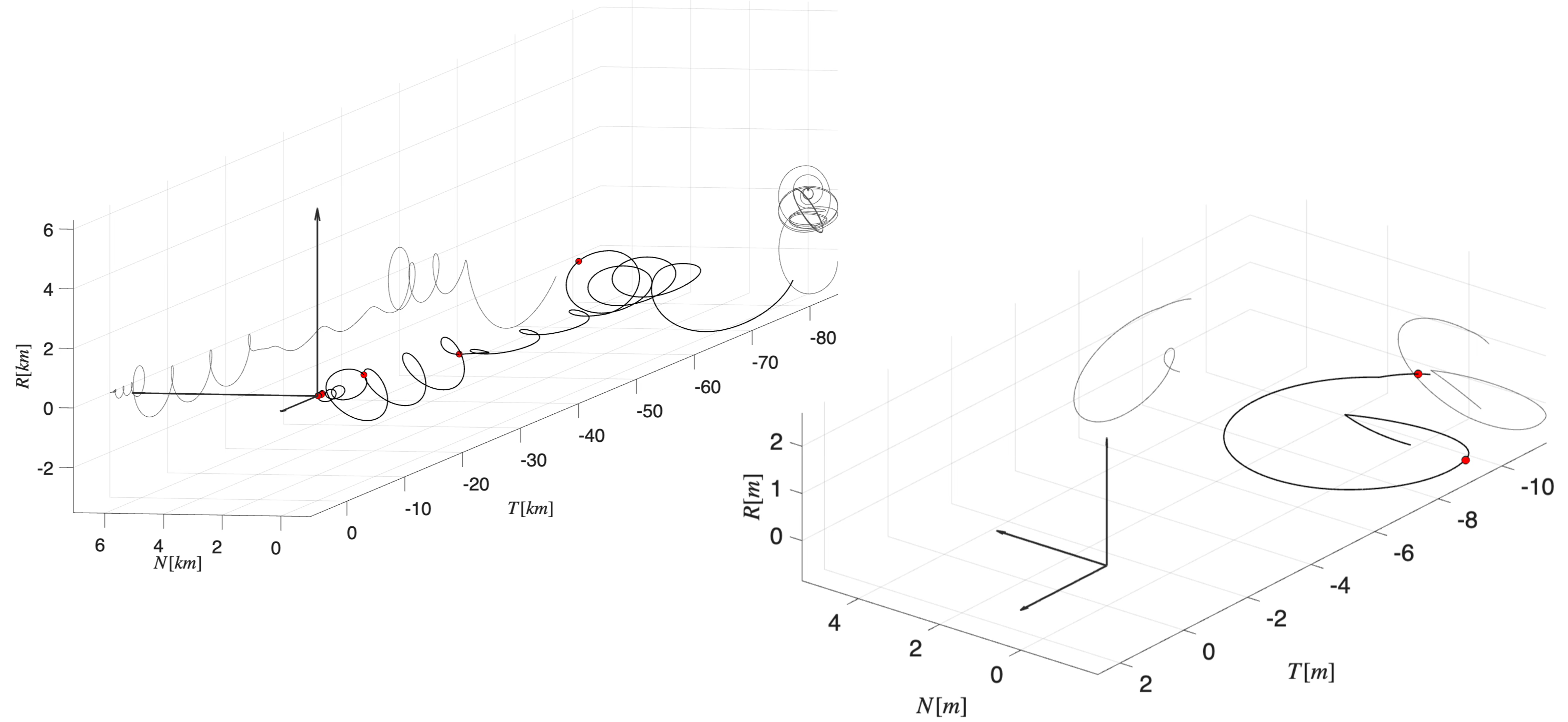}
    \caption{Example far-range (top left) and close-range (bottom right) trajectories that target waypoints 0-7 and waypoints 8-10, respectively. The individual waypoints are shown by the red circles and the $T$-$R$ and $N$-$R$ projections are shown projected in grey.}
    \label{fig:sample-trajectories}
\end{figure}

Next, the servicer begins a V-bar approach to an along-track separation of 10 m (waypoint \#8), where the target becomes fully resolved in the camera frame. 
At waypoint \#9, the servicer transfers to a passively safe relative e-/i- vector separated orbit at an along-track distance of 7 m.
Finally, at waypoint \#10, the servicer maneuvers to reduce the size of the relative orbit to a radius of 1 m and retreat to a separation of 8.5 m.
An example of the far- and close- range portions of the trajectory can be seen in \autoref{fig:sample-trajectories}. These trajectories are generated using ground truth state knowledge, i.e. ideal navigation, while including control to achieve the desired waypoints. The far-range trajectory is computed using the Chernick D'Amico Closed-Form control solver\cite{chernick_new_2018,chernick_closed-form_2021}, while the close-range trajectory uses the Koenig-D'Amico controller \cite{koenig2020fast}. This is consistent with the controllers employed in each experiment, as shown in Table~\ref{tab:experiments_table} and discussed further in the following section.

\subsection{Experiment Specifications}

\begin{table}[t]
    \footnotesize
    \centering
    \caption{Specifications for each experiment in this work.}
        \begin{tabular}{rcccc}
        
        \toprule
        \textbf{Experiment} & \textbf{1} & \textbf{2} & \textbf{3} \\
        \midrule
        \textbf{ Range} & Far & Close & Close \\
        \midrule
        \makecell[r]{\textbf{Waypoints} \\ (Table~\ref{tab:full range waypoints})} & 0-7 & 8-10 & 8-10 \\ 
        \midrule
        \makecell[r]{\textbf{Cooperative} \\ \textbf{Target?}} & No & No  & Yes \\ 
        \midrule
        \makecell[r]{\textbf{Servicer}\\ \textbf{Attitude}} & 
        \makecell{Camera boresight \\ pointed at \\ target center } & \makecell{Camera boresight \\ pointed at \\ target center} & \makecell{Zenith \\ pointing} \\ 
        \midrule
        \makecell[r]{\textbf{Target}\\ \textbf{Attitude}}  & Tumbling & Tumbling & \makecell{Zenith \\ pointing} \\ 
        \midrule
        \makecell[r]{\textbf{Navigation} \\ \textbf{Algorithm(s)}}  & AOT + SEDF & SPN + SEDF & CDGNSS + IAR + SEDF \\
        \midrule
        \makecell[r]{\textbf{Control}\\ \textbf{Algorithm(s)}} & Closed-Form & \makecell{Koenig-D'Amico \\ Least Squares (Backup)} & \makecell{Koenig-D'Amico \\ Least Squares (Backup)} \\ 
        \midrule
        \makecell[r]{\textbf{SIL Sensor}\\\textbf{Model}} & \makecell[c]{OpenGL \\ renderer} & \makecell[c]{OpenGL \\ renderer} & \makecell[c]{GNSS Software \\ Emulator} \\
        \midrule
        \textbf{HIL Testbed} & OS & TRON & GRAND \\
        \midrule
        \makecell[r]{\textbf{Measurement} \\ \textbf{Rate (Hz)}} & 0.0167 & 0.2 & 0.1 to 1 \\
    \toprule
    \end{tabular}
    \label{tab:experiments_table}
\end{table}

Table \ref{tab:experiments_table} summarizes the testing configuration for three closed-loop experiments.
Experiments are categorized as close-range or far-range depending on the largest ISD associated with the trajectory waypoints used.

Waypoints 0-4 correspond to far-range separations on the order of kilometers, where the target spacecraft is fully unresolved. Waypoints 5-7 correspond with mid-range separations, where the target is semi-resolved. Finally, waypoints 8-10 correspond to close-range separations below 10 m, where the target is fully resolved in imagery. 
In each experiment, the first waypoint in its range (i.e., waypoint 0 or 8) is used as the initial condition.
Experiments 1 and 2 assume the target is uncooperative, meaning that it does not share navigation information with the servicer spacecraft, and they assume its attitude is not controlled.
In contrast, Experiment 3 uses a cooperative target, meaning that it shares navigation information with the servicer via inter-satellite link, and it is assumed that both satellites' attitudes are controlled to maintain the RF crosslink.

As indicated in Table \ref{tab:experiments_table}, navigation and control algorithms for each experiment are selected based on the mission profile. Experiments 2 and 3 use the Koenig-D'Amico control solver as the default. 
However, if the convex optimizer does not converge to a solution, the Least Squares solver is used as backup. 
For all experiments, actuators are modeled ideally, with commanded $\Delta V$ is applied to the servicer satellite without error.
Each experiment is conducted twice, once in the SIL testing mode and once using the HIL testbeds.

\section{Results}
\label{sec:results}

This section describes the SIL and HIL testing results from each experiment.
The GNC system's performance is evaluated using two metrics: navigation error and control error taken at each waypoint time.
Navigation error is the difference between the estimated and the ground truth RTN position of the target, while control error represents the deviation between the commanded and ground-truth waypoints in RTN space.
Both metrics are presented as the L2-norm of the position components.
In addition, the navigation algorithm in Experiment \#2 estimates the full target pose (i.e. position and attitude); therefore, the attitude error is also reported. This metric is defined as the total angular error between the true and estimated relative attitudes, where relative attitude is the target's attitude expressed in the camera frame.

\subsection{Experiment \#1 Results: Uncooperative Far-Range using Angles-Only Tracking (AOT)}

\subsubsection{Navigation Performance:}

\begin{figure}[h]
    \centering
    \includegraphics[width=\linewidth]{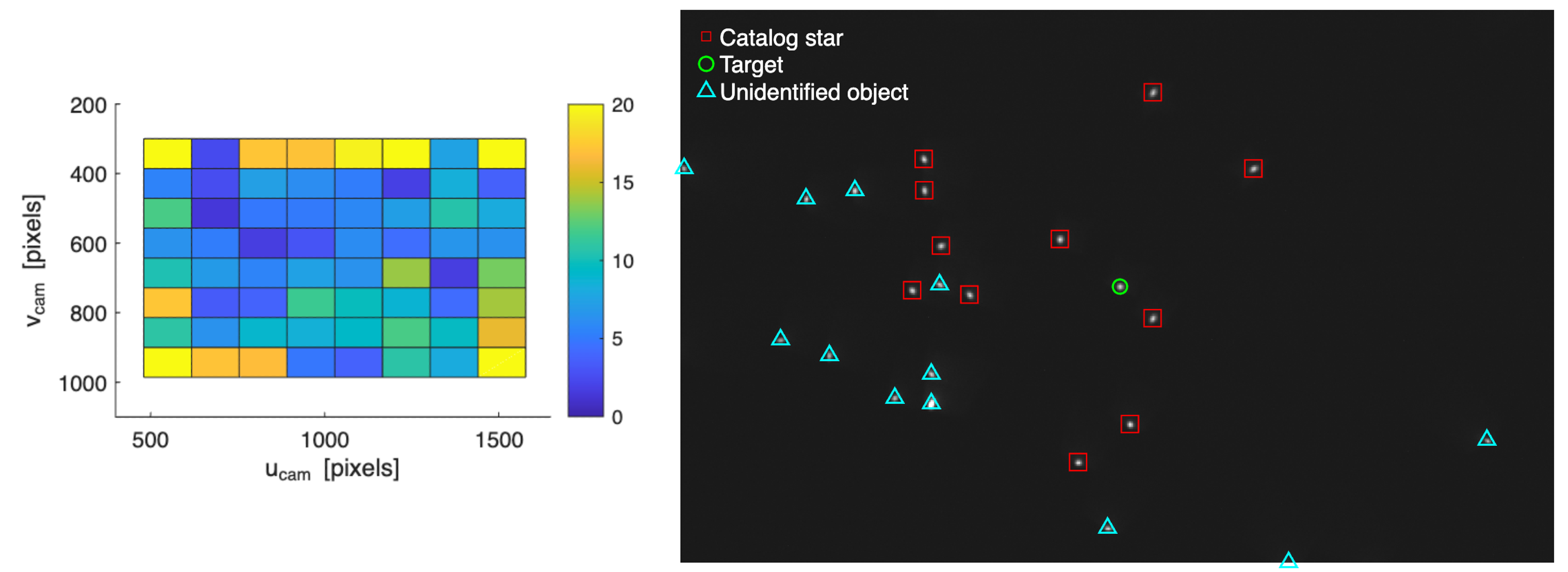}
    \caption{(\emph{Left}) Optical Stimulator calibration residuals for the central region-of-interest. Residuals are shown as angular error in arcseconds between the known unit vectors associated with each calibration point and the corresponding unit vectors recovered from VBS measurements after distortion correction. The heatmap illustrates the spatial distribution of residuals in the sensor's image plane.(\emph{Right}) Sample HIL image used for AOT, showing point sources detected by the Image Processing (IMP) module. Identified catalog stars are denoted by red boxes, the target spacecraft is denoted by a green circle, and objects detected by the IMP module that do not match any catalog entries are denoted by blue triangles.}
    \label{fig:artms_ex_sil_img16}
\end{figure}

Both the SIL and HIL tests for Experiment \#1 use synthetically rendered images, with the difference that the HIL test captures the image using real hardware mounted on the OS while the SIL image is ``captured'' using a camera model. Thus, the difference between the SIL and HIL results isolates the effect of the true optics that are not captured by the ideal camera model. To limit the impact of residual calibration errors in the OS, the image processing module used by AOT is restricted to the central 500 x 800 pixel region-of-interest (ROI), where the testbed calibration yielded geometric residuals at or below 10 arcseconds, as shown in Figure~\ref{fig:artms_ex_sil_img16}. This ROI is imposed for both the SIL and HIL tests. Figure~\ref{fig:artms_ex_sil_img16} also shows an example of a HIL image used for AOT with the detected point sources. 

Table~\ref{tab: far-range table (ARTMS)} includes the RTN navigation errors from the SIL and HIL tests for Waypoints 1-7 from Table~\ref{tab:full range waypoints}. Notably, both SIL and HIL exhibit similar navigation errors across the trajectory. This result indicates that within the calibrated ROI, optical and hardware-induced effects are not a limiting factor in the AOT algorithm's ability to process the incoming measurements. 

\begin{table}[b!]
\centering
\caption{ Performance Metrics for Experiment \#1 (far-range uncooperative approach). Navigation and control error are expressed as the L2 norms of the RTN position at each waypoint.}
\begin{tabular}{c|c|rrrrrrr}
& Waypoint & 1 & 2 & 3 & 4 & 5 & 6 & 7\\
\hline
\multirow{2}{*}{\textbf{Navigation Error (m)}} & SIL & 80.04 & 50.75 &3.18 & 2.57 & 4.57 & 1.33 & 3.89 \\
& HIL & 75.27 &23.03 & 9.58 & 6.70 & 10.12 & 1.09 & 3.06 \\
\hline
\multirow{2}{*}{\textbf{Control Error (m)}} & SIL & 24.40 &10.26 &2.47 &3.91 &8.01 &3.64 &8.77 \\
& HIL & 113.88 &73.59 &13.17 &6.44 &10.11 &1.22 &4.47  \\
\hline
\end{tabular}
\label{tab: far-range table (ARTMS)} 
\end{table}

\begin{figure}[ht]
    \centering
    \includegraphics[width=\linewidth]{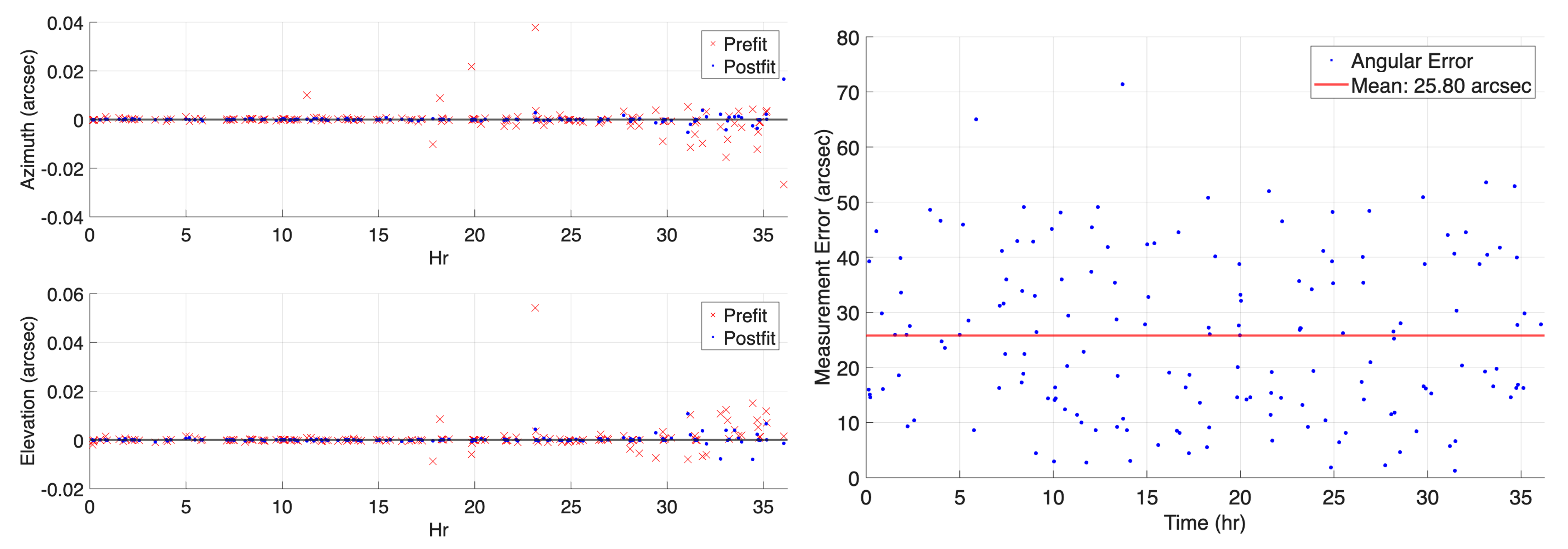}
    \caption{(\emph{Left}) SIL Navigation filter pre- and post-fit residuals for AON bearing angles over the course of the far-range trajectory. (\emph{Right}) SIL Angular error between the ground truth and measured bearing angles. All values are provided in arcseconds.}
    \label{fig:artms_residuals_measurement_errors}
\end{figure}

While the navigation error decreases as the relative spacecraft separation decreases, this reduction is not proportional to the change in the separation. In fact, the navigation error increases relative to the separation, growing from approximately 0.16\% of the target range at waypoint \#1 to approximately 3\% at waypoint \#7. 
Evaluating the navigation filter pre-fit residuals over the trajectory reveals that the residuals increase at later waypoints as the separation decreases, consistent with the observed growth in relative navigation error, while the measurement errors remain approximately constant. The filter residuals and measurement errors can be seen in Figure~\ref{fig:artms_residuals_measurement_errors} in \nameref{sec:appendix_2}.
Since the angular measurement error remains approximately constant, the growth in the relative navigation error can be attributed to a reduction in target observability with constant maneuver frequency at closer separations. At waypoints \#5 through \#7, the size of the relative orbit decreases significantly, resulting in reduced out-of-plane motion and therefore diminished observability of the state. This leads to an increase in the navigation errors relative to separation. Despite this effect, estimation of all of the ROE remains robust overall, and is consistent between SIL and HIL.

\subsubsection{Control Performance:}

\begin{figure}[t]
    \centering
    \includegraphics[width=1\linewidth]{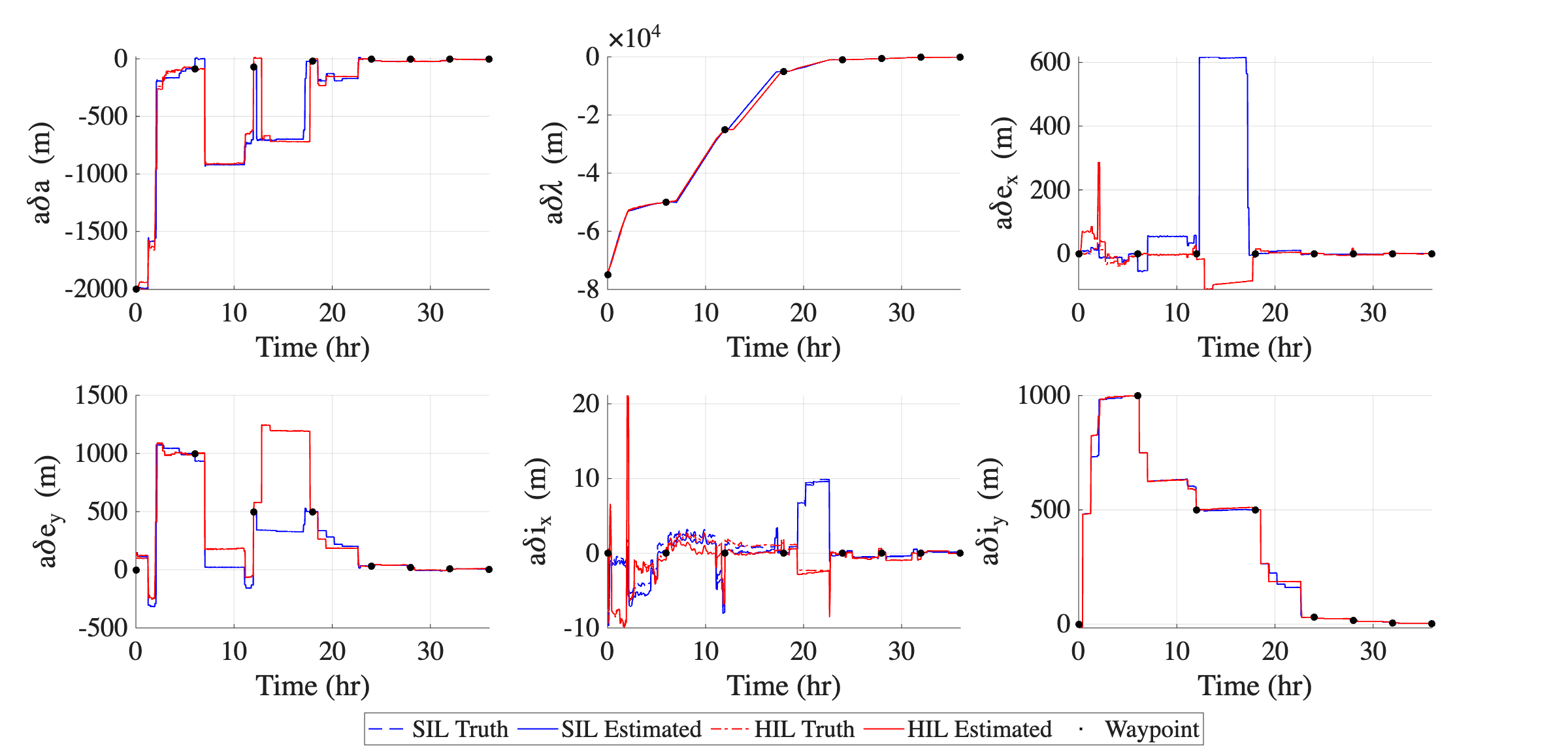}
    \caption{Results of Experiment \#1: Actual, estimated, and desired ROE of the target from SIL and HIL testing. ROE are plotted as mean quantities.}
    \label{fig:far-range-roe-traj}
\end{figure}

\begin{figure}[ht]
    \centering
    \includegraphics[width=0.75\linewidth]{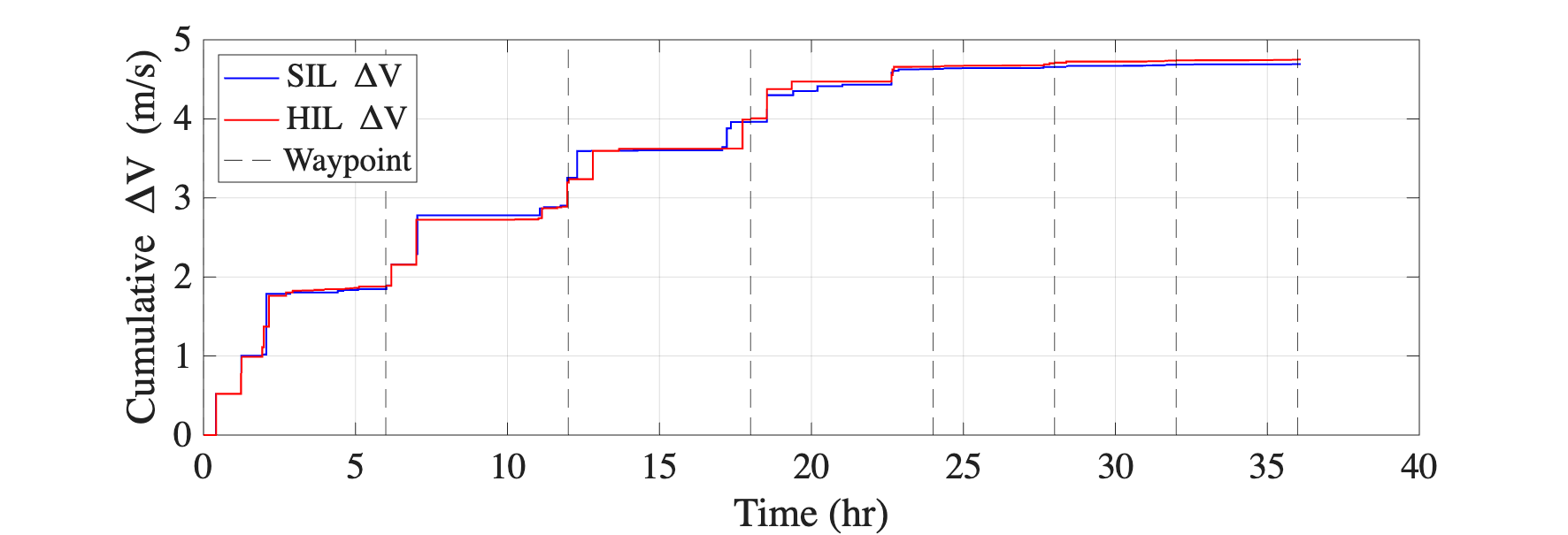}
    \caption{SIL and HIL cumulative $\Delta V$ during Experiment \#1 using AOT. In total, SIL executed 61 maneuvers (\SI{4.69} {m/\second}) and HIL executed 90 maneuvers (\SI{4.75} {m/\second}). The dashed vertical lines indicate the times associated with each waypoint in the trajectory.}
    \label{fig:artms_results_deltav_cumulative}
\end{figure}

With respect to the control errors, commanded maneuvers, and resultant trajectory, the most notable difference between SIL and HIL is the change in $a\delta e_x$ and $a\delta e_y$ between waypoints \#2 and \#3, as shown in Figure~\ref{fig:far-range-roe-traj}. The maneuver timings in Figure~\ref{fig:artms_results_deltav_cumulative} reveal that GNC in the SIL test executes a maneuver immediately after waypoint \#2 at 12 hours, whereas HIL performs a maneuver of the same magnitude with a slight delay. Because the relative position differs at the two maneuver times, the SIL case exhibits a larger change in the x-component of the relative eccentricity, while the HIL case exhibits a larger change in the y-component. Because the GNC software is deterministic, the difference in the maneuver timings can be traced to the small discrepancy in the navigation errors at waypoint \#2. Nevertheless, both SIL and HIL produce valid trajectories with similar cumulative $\Delta V$ values (4.69 vs. 4.75 m/s, respectively) as well as similar control errors at each waypoint as shown in Table~\ref{tab: far-range table (ARTMS)}.



\subsection{Experiment \#2 Results: Uncooperative Close-Range using Spacecraft Pose Network (SPN)}

\subsubsection{Navigation performance:}

\begin{figure}[b!]
    \centering
    \includegraphics[width=\linewidth]{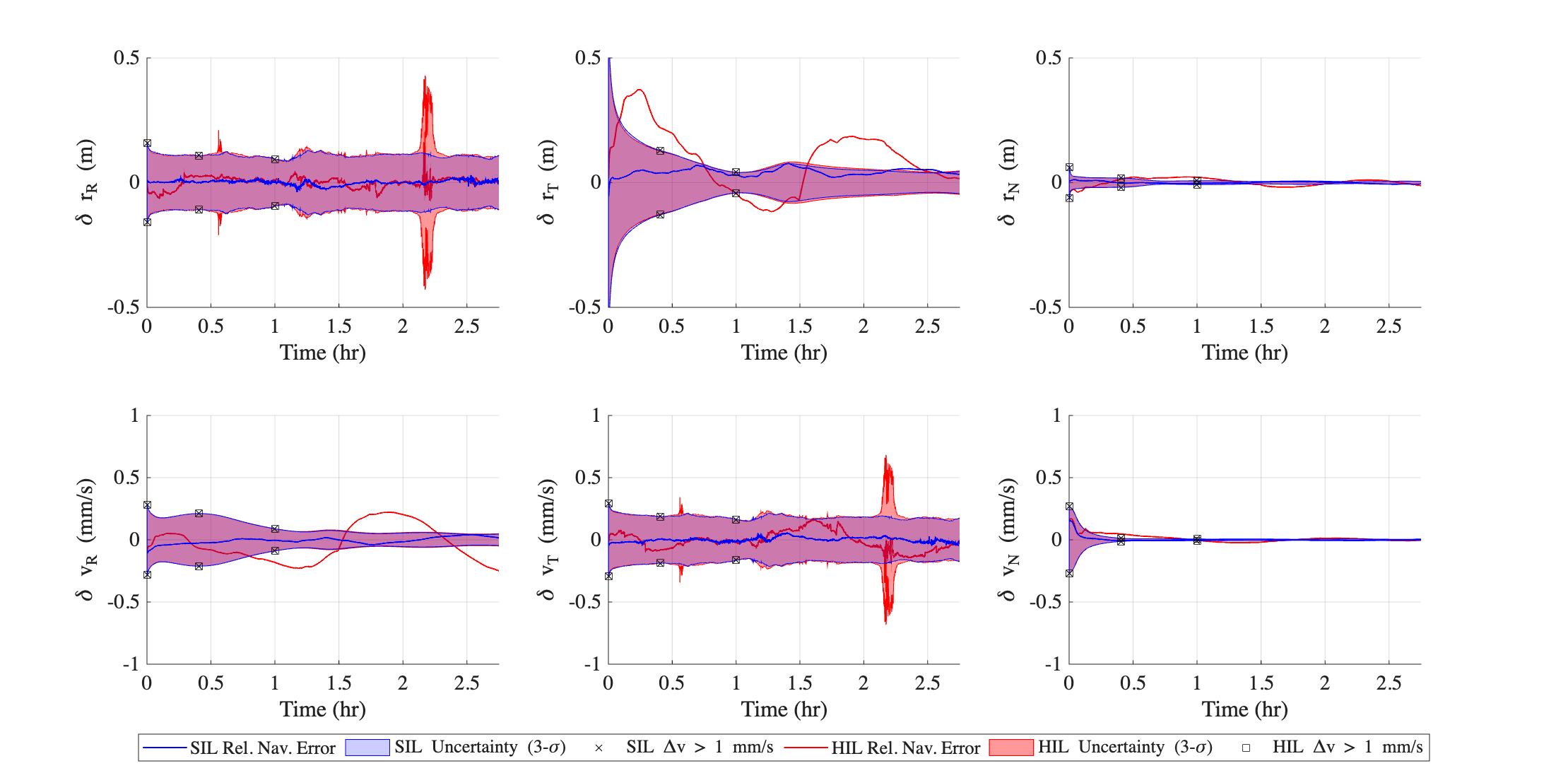}
    \caption{Navigation performance of Experiment \#2: True error and filter covariances of the relative positions (top) and velocities (bottom), expressed in the RTN frame, from SIL and HIL testing. Large maneuvers are also plotted on the corresponding covariance. }
    \label{fig:spn_results_roe_nav}
\end{figure}

\begin{figure}[t!]
    \centering
    \includegraphics[width=\linewidth]{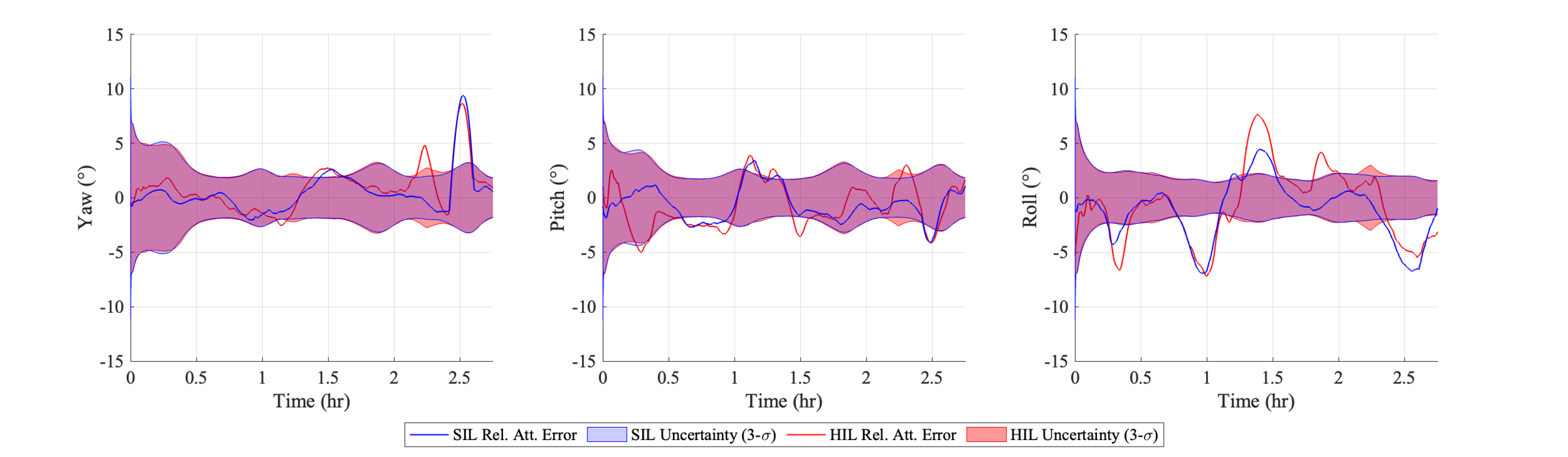}
    \caption{Navigation performance of Experiment \#2: True error and filter covariance of the relative attitude of the target from SIL and HIL testing. }
    \label{fig:spn_results_target_att_nav}
\end{figure}

As shown in Table \ref{tab: close range table (SPN)}, the GNC achieves position navigation accuracy on the order of centimeters at both waypoints. 
The HIL test showed higher error at both waypoints, suggesting that while the SIL and HIL experiments created similar inputs to the GNC stack, the HIL images are more challenging for the SPN module than the rendered images.
This outcome matches expectations, as SPN was trained using synthetic imagery and its performance is expected to degrade in HIL due to the domain gap between synthetic and hardware-based images\cite{park_robust_2023}.
\autoref{fig:spn_results_roe_nav} shows the navigation filter performance using SPN measurements, with all maneuvers exceeding 1 mm/s indicated. 
As expected from \autoref{tab: close range table (SPN)}, the filter's estimate of the state is worse in HIL testing than for the SIL experiment. 
Notably, the filter struggles to estimate along-track separation in the HIL testing, which is unsurprising as this is most closely aligned with the camera boresight direction and is therefore the least observable for a VBS. 
Also shown in Table \ref{tab: close range table (SPN)}, relative attitude error is at the order of single degrees throughout the trajectory, though performance suffers slightly more in the HIL case.
The UKF's attitude estimation performance over time is shown in \autoref{fig:spn_results_target_att_nav}.

\begin{table}[ht]
\centering
\caption{Performance metrics for Experiment \#2 (uncooperative close-range rendezvous). L2 norms of RTN position error, relative target attitude error, and RTN position error for control are given at each waypoint time. Navigation error mean and standard deviation are computed over the experiment duration.}
\begin{tabular}{c|c|rr|r}
& Waypoint & 9 & 10 & Mean ± Std\\
\hline
\multirow{2}{*}{\textbf{Navigation Error (m)}} & SIL & 0.0350 & 0.0512 & 0.0457 $\pm$ 0.0143 \\
& HIL & 0.0572 & 0.1035 & 0.1250 $\pm$ 0.0956 \\
\hline 
\textbf{Relative Attitude} & SIL & \multicolumn{2}{ c |}{-} & 3.1615 $\pm$ 2.3551 \\
\textbf{Navigation Error ($^\circ$)} & HIL & \multicolumn{2}{ c |}{-} & 4.1302 $\pm$ 2.1608  \\
\hline
\multirow{2}{*}{\textbf{Control Error (m)}} & SIL & 0.2941 & 0.1231  & \multicolumn{1}{c}{-} \\
 & HIL & 0.0898 & 0.2450  & \multicolumn{1}{c}{-} \\
\hline
\end{tabular}
\label{tab: close range table (SPN)}
\end{table}

The filter performance for both relative orbit and pose show a similar trend: throughout most of the trajectory, the state covariance is nearly identical between the SIL and HIL tests, demonstrating high consistency between the software- and hardware-based GNC inputs.
However, at approximately 2.25  hours, the state covariance in the HIL test shows a brief  ``jump" in the $a \delta a$ ROE component and all attitude components. 
A similar trend is seen in PnP pose errors across the experiment, shown in Figure \ref{fig:spn_results_spn_measurement_errors} in Appendix B, indicating that SPN computed poor estimates of keypoint locations for a brief window.
The filter responds as expected by increasing the covariance to account for the mis-match between the measurements from SPN and the modeled measurement based on the current state estimate. 

\begin{figure}[htb!]
    \captionsetup[subfigure]{labelformat=empty} 
    \centering
    \begin{subfigure}{0.49\textwidth}
        \includegraphics[width=\textwidth]{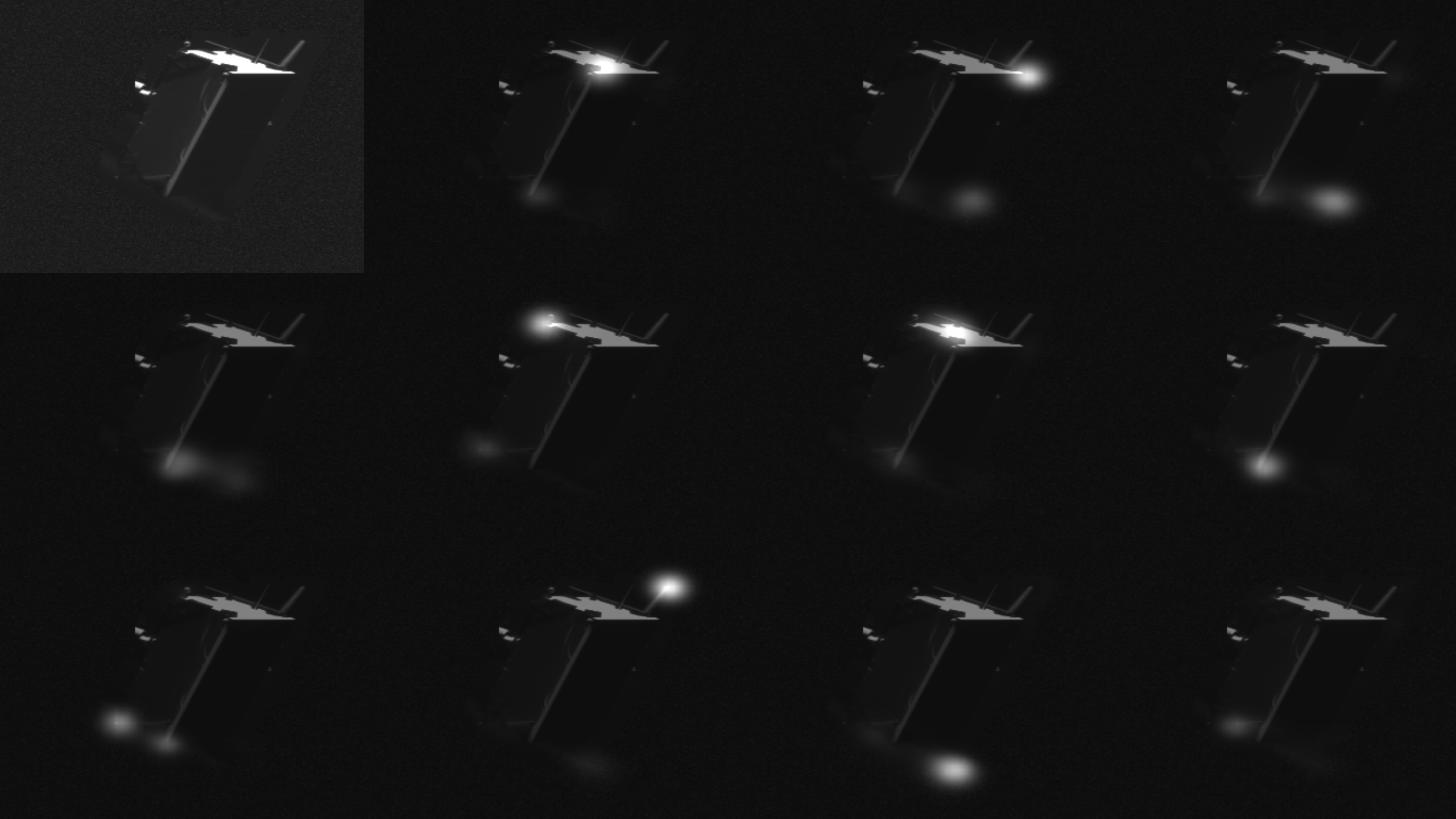}
    \end{subfigure}
    \begin{subfigure}{0.49\textwidth}
        \includegraphics[width=\textwidth]{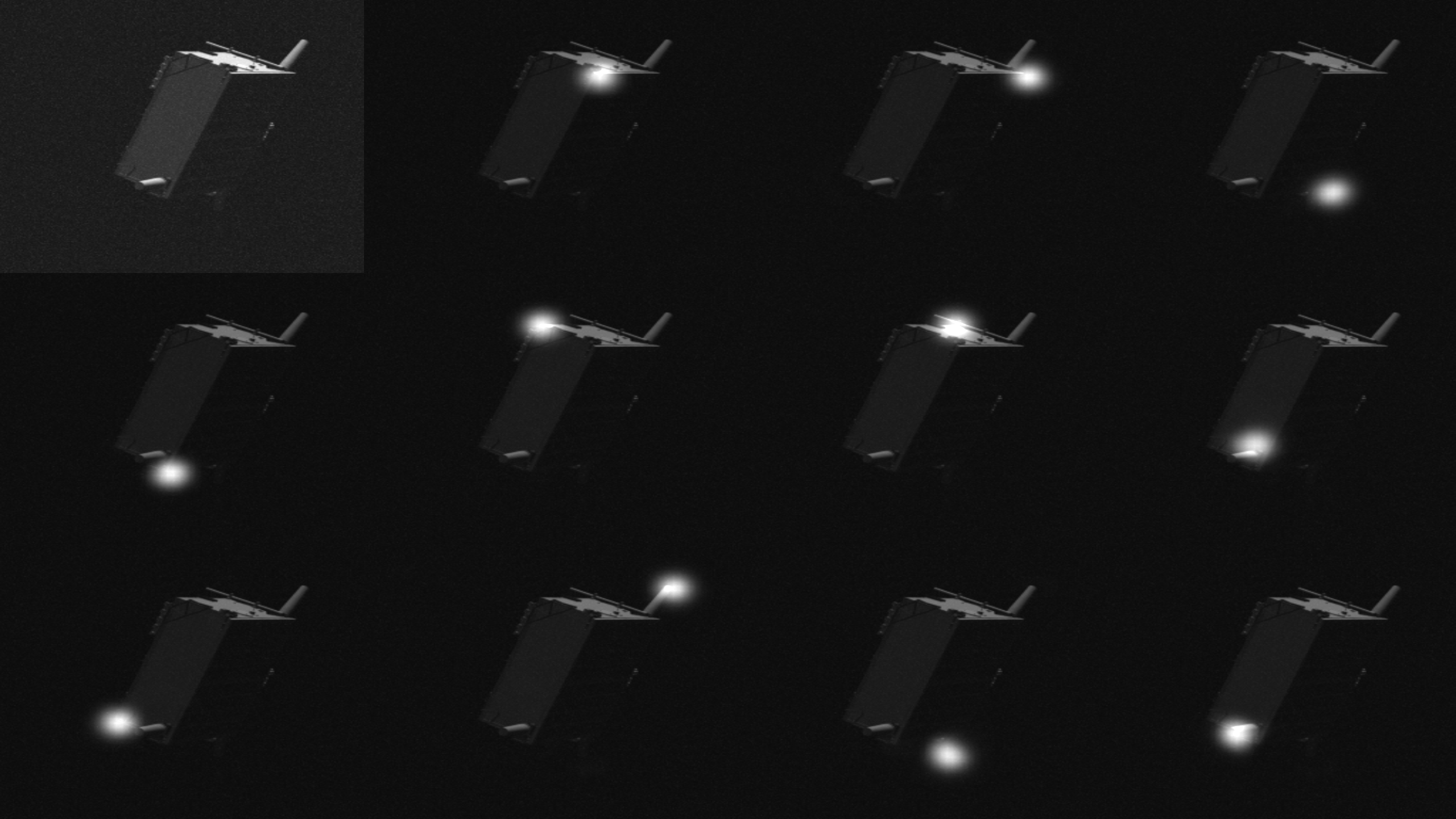}
    \end{subfigure}
    \caption{Example of HIL SPN outputs during the period of high measurement error seen in \autoref{fig:spn_results_spn_measurement_errors} after hour 2. Left side shows a HIL image from this period (upper left corner), followed by images of SPN's ``heatmap" estimates of keypoint locations overlaid on the input image.
    Right side shows a re-creation of the same pose in OpenGL based renderer, along with SPN outputs for rendered image.}
    \label{fig:spn_example_images}
\end{figure}

The cause of this period of poor measurement performance can be traced back to the imagery itself.
Figure \ref{fig:spn_example_images} shows an example of SPN's outputs when given the HIL image associated with the highest PnP pose error, alongside SPN's outputs given a rendered image of the same scene.
Due to minute  lighting and texture differences between SIL and HIL, the HIL image shows reduced contrast between various target components and the background, making detection more challenging.
Furthermore, an artifact from the HIL testbed is visible to the left of the spacecraft due to imperfect application of the post-processing mask. 
Both of these factors lead to poor-quality estimates of keypoint locations, as demonstrated in Figure \ref{fig:spn_example_images}.
SPN's keypoint estimates are initially provided as heatmaps, which are then post-processed to provide estimated keypoint locations and keypoint measurement covariances. \cite{park_robust_2023}
The heatmaps for the HIL image in Figure \ref{fig:spn_example_images} show many faint, spread out, or bimodal distributions; in contrast, all heatmaps for the rendered image are unimodal and distinct.
Because measurement covariance is computed from the heatmaps themselves, the spread out distributions in the HIL image likely contributed to the elevated state covariance at around 2.1 hours.



Finally, it must be noted that a portion of navigation error in the HIL tests is introduced by the testbed itself.
TRON's robotic arms achieve the commanded poses with some small amount of translation and rotation error; this error will cause the SPN pose measurement to diverge slightly from the true pose.
However, as shown in Table \ref{tbl:rwhe_results}, the TRON facility's translation and rotation errors are at the millimeter and sub-degree scale respectively.
Because errors of this magnitude are miniscule compared to the navigation error from the GNC stack as seen in the SIL test, they are neglected in this work.

\subsubsection{Control performance:}

\begin{figure}[hb!]
    \centering
    \includegraphics[width=0.95\linewidth]{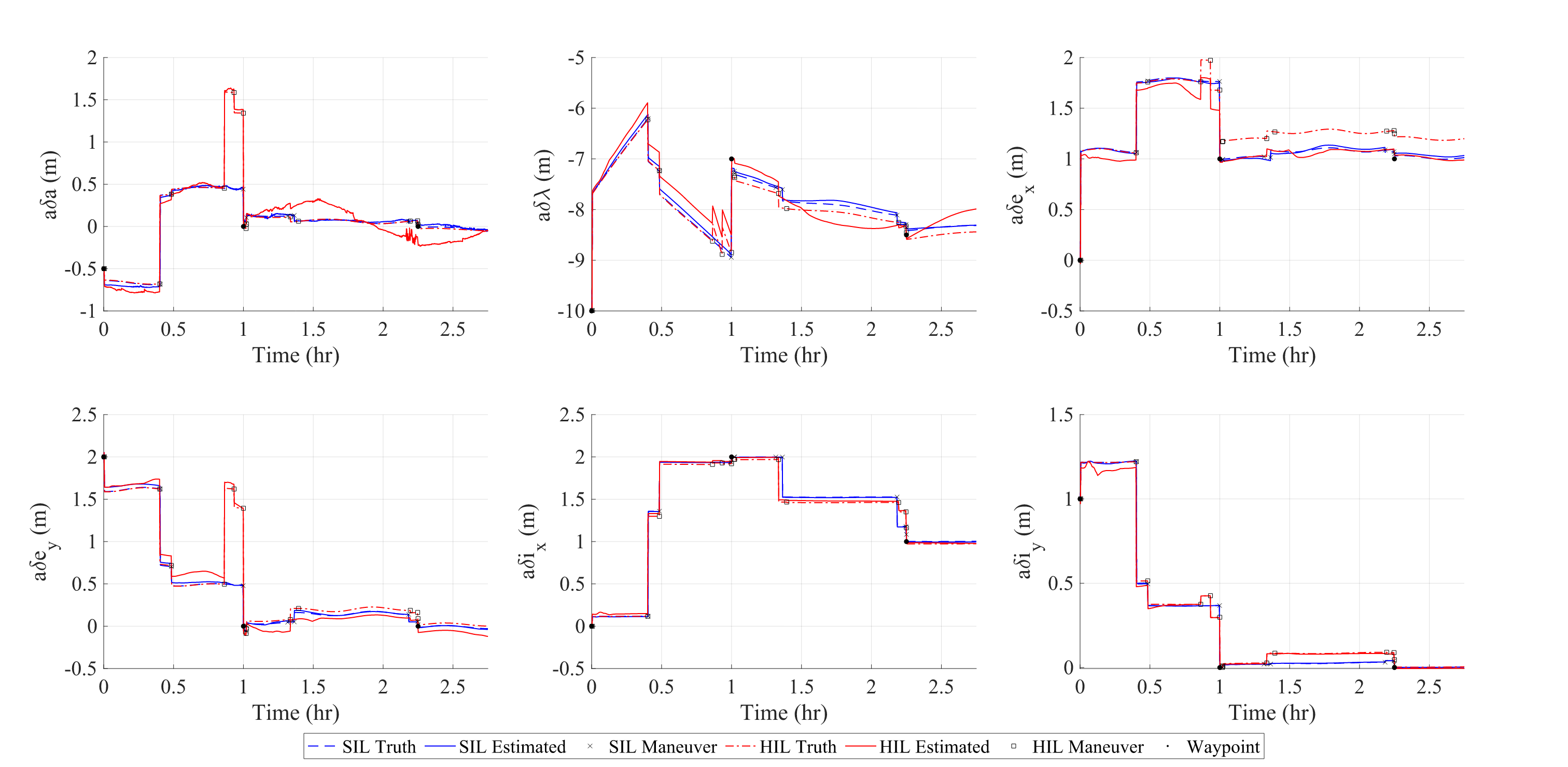}
    \caption{Results of Experiment \#2: Actual, estimated, and desired ROE of the target from SIL and HIL testing. ROE are plotted as osculating quantities with impulsive maneuvers plotted on top.}
    \label{fig:spn_results_roe_man}
\end{figure}

\begin{figure}[H]
    \centering
    \vspace{-3mm}
    \includegraphics[width=0.9\linewidth]{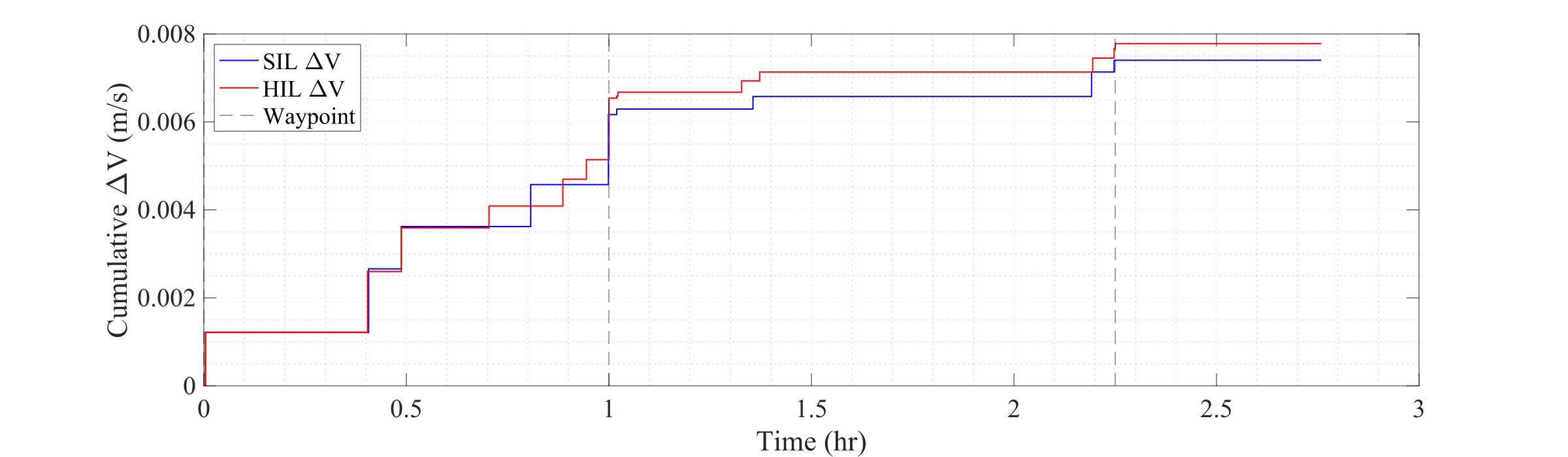}
     \caption{SIL and HIL cumulative $\Delta V$ during Experiment \#2. In total, SIL executed 11 maneuvers (\SI{5.78} {mm/\second}) and HIL executed 13 maneuvers (\SI{6.91} {mm/\second}) (20\% increase from HIL over SIL). The dashed vertical lines indicate the time associated with each waypoint in the trajectory.}
    \label{fig:spn_results_deltav_cumulative}
\end{figure}

The true and estimated trajectories of the spacecraft throughout the SIL and HIL experiments are shown in Figure \ref{fig:spn_results_roe_man}. 
Two trends are evident in these plots.
Due to increased estimation error in the HIL test, there is a visible offset between the estimted and true HIL trajectory in many of the ROE components.
In addition, the HIL trajectory initiates a cluster of maneuvers shortly before the 1 hour mark, leading to a brief yet significant departure of the HIL trajectory from the SIL trajectory.
This cluster of maneuvers was initiated by the least squares control solver, which was activated when the estimated ROE error grew beyond the allowable threshold.
Finally, $\Delta V$ over the course of each experiment is shown in Figure \ref{fig:spn_results_deltav_cumulative}. 
The GNC stack makes similar maneuver commands between the SIL and HIL experiments.
However, the initial cluster of manuevers leads greater $\Delta V$ expenditure in HIL than SIL tests.

\subsection{Experiment \#3 Results: Cooperative Close-Range using CDGNSS with IAR}

\subsubsection{Navigation performance:}

\begin{figure}[htb]
  \centering
  \begin{subfigure}[t]{0.49\textwidth}
    \centering
    \includegraphics[width=\linewidth]{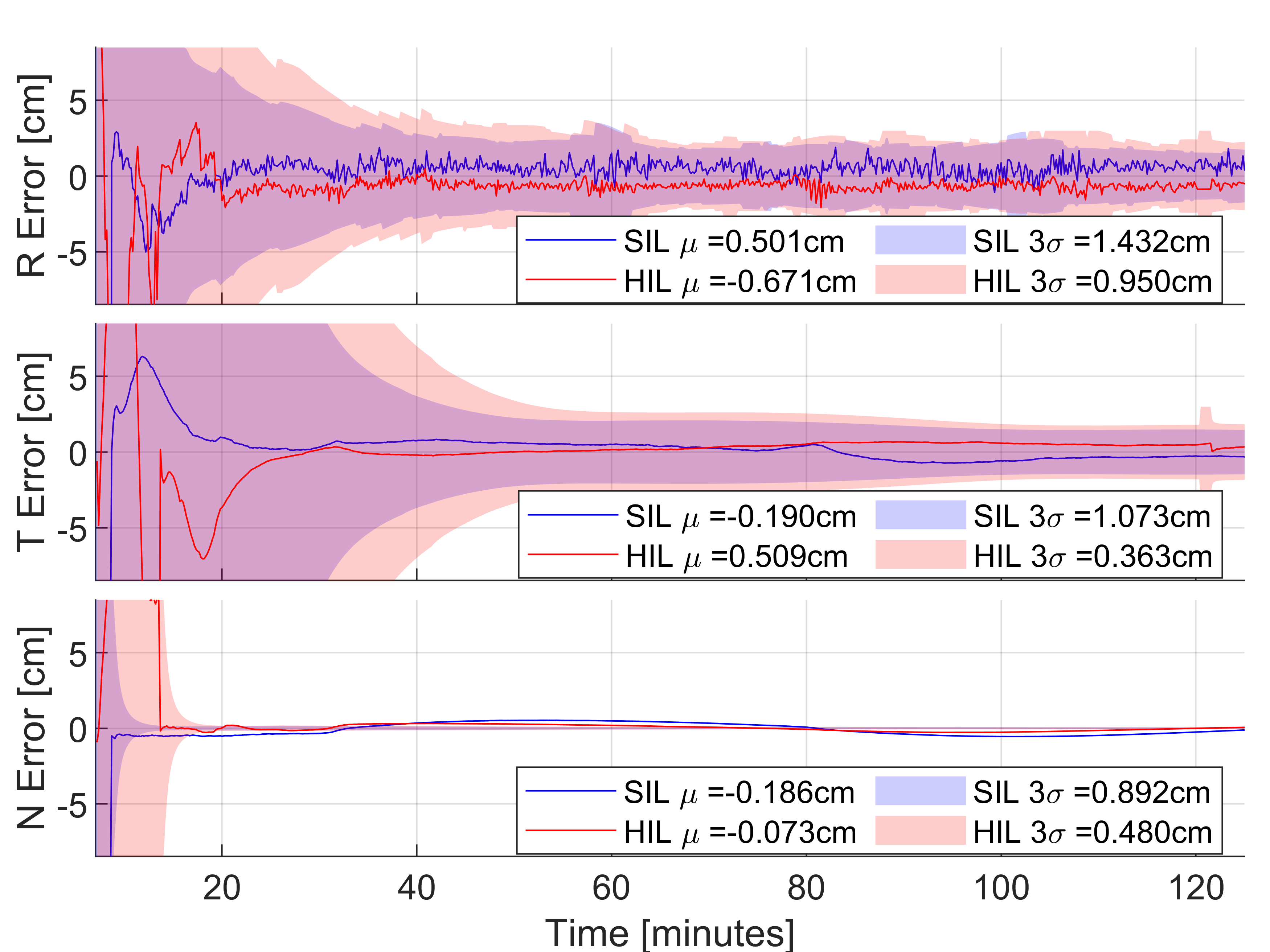}
  \end{subfigure}
  \hfill
  \begin{subfigure}[t]{0.49\textwidth}
    \centering
    \includegraphics[width=\linewidth]{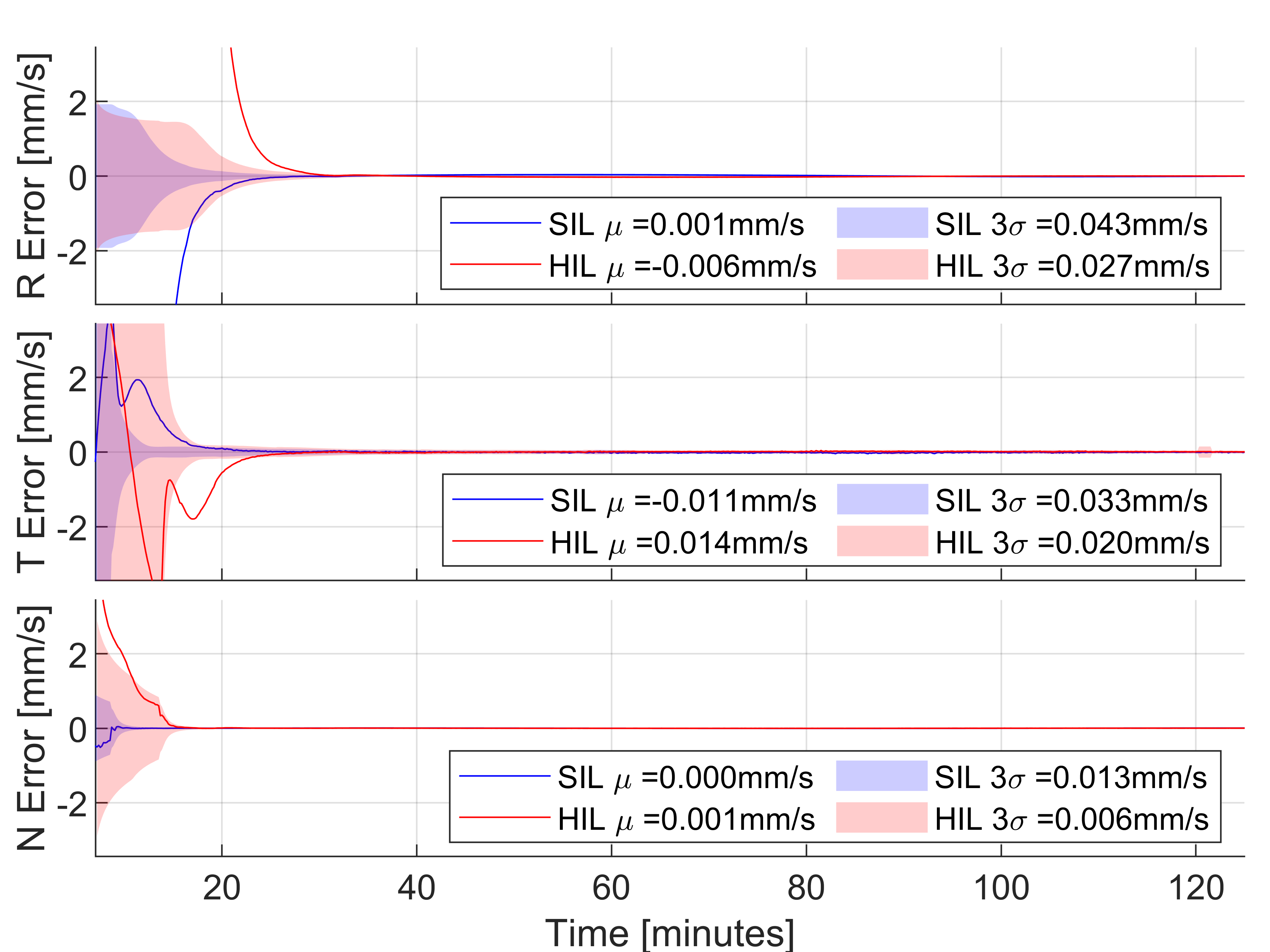}
  \end{subfigure}
  \caption{Navigation performance of Experiment \#3: True error and filter covariances of the relative positions (left) and velocities (right), expressed in the RTN frame, from SIL and HIL testing.}
  \label{fig:dgtl-rtn-err}
\end{figure}

The CDGNSS with IAR module \cite{giralo2019digital, low2026digital} embedded in the GNC stack \cite{Kruger2024} is evaluated under both SIL with the GNSS Software Emulator and HIL with GRAND.
Both SIL and HIL have the capability of emulating a GNSS receiver under cold-start (no prior almanac) or warm-start (with prior almanac) conditions.
In this experiment, the warm-start condition is applied to both.
Upon reaching Waypoint 8, CDGNSS measurements commence. \autoref{fig:dgtl-rtn-err} presents the relative position and velocity navigation errors in the RTN frame for both cases.
Under the same modeled conditions for initialization and measurement variance, the steady-state errors under SIL and HIL conditions are similar and consistent. However, the convergence of state errors under a HIL measurement source are more gradual.
This stems from non-stationary processes arising from actual receiver hardware tracking-loop dynamics and temporal correlations, which introduces differences between SIL- and HIL-generated measurements.
These are challenging to capture in full fidelity under SIL conditions.

\subsubsection{Control performance:}

\begin{figure}[htb]
    \centering
    \includegraphics[width=0.85\linewidth]{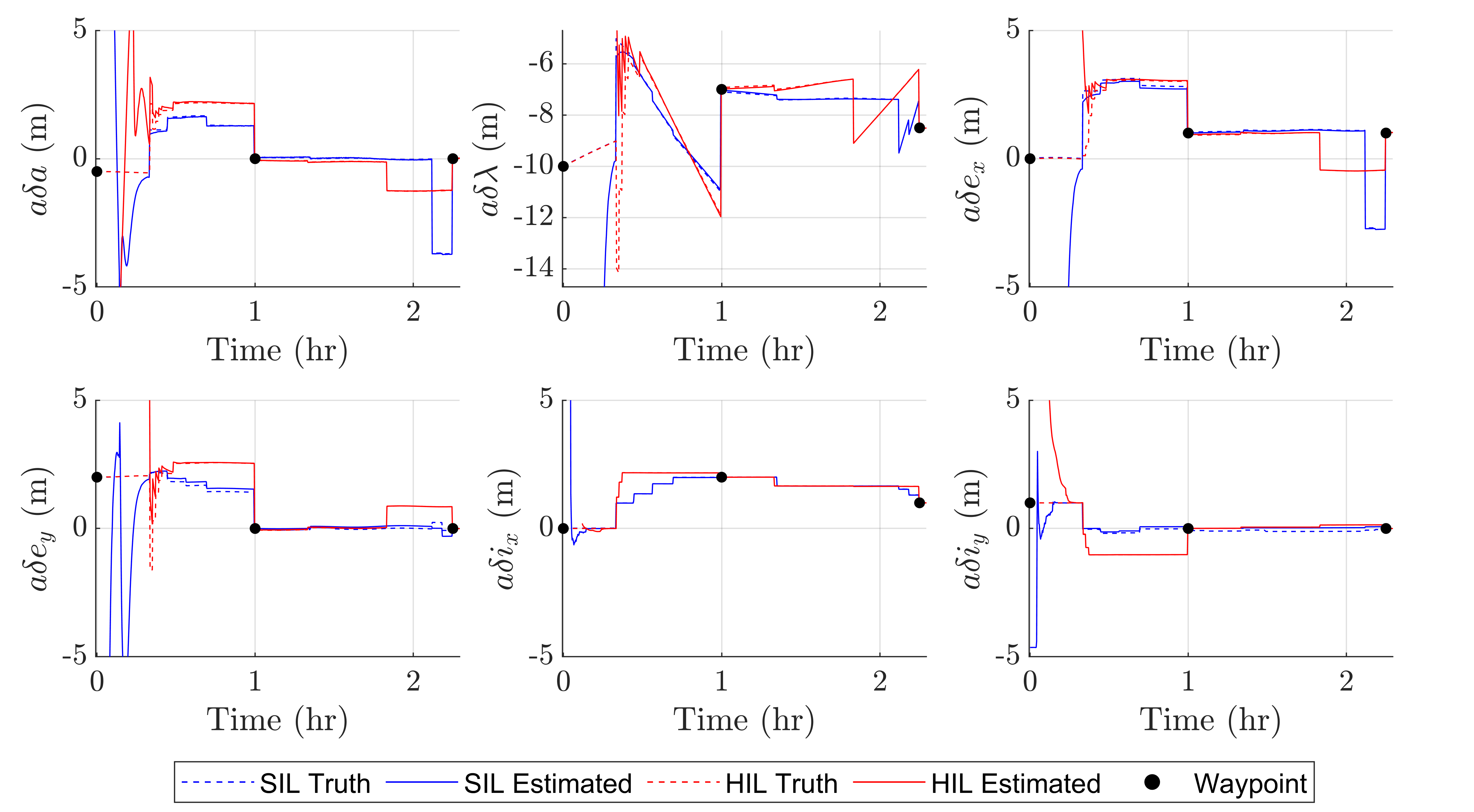}
    \caption{Results of Experiment \#3 using CDGNSS with IAR: Actual, estimated, and desired ROE of the target from SIL and HIL testing. ROE are plotted as osculating quantities.}
    \label{fig:dgtl-roe-waypoints}
\end{figure}

In this experiment, the maneuver planner was initiated at $t = 20$ minutes, to allow time for filter convergence.
The trajectories in ROE space shown in \autoref{fig:dgtl-roe-waypoints} indicate that the majority of commanded waypoints are successfully reached in both configurations with mm-level navigation errors and sub-meter waypoint control errors, tabulated in \autoref{tab:close-range-results-dgtl}.

\begin{table}[htb]
\centering
\caption{Performance metrics for Experiment \#3 (cooperative close-range) using the CDGNSS with IAR \cite{low2026digital}. L2 norms of navigation and control errors are given at each waypoint time, while navigation error statistics are computed for the whole trajectory post-IAR.}
\begin{tabular}{c|c|rr|r}
 & Waypoint & 9 & 10 & Mean $\pm$ Std \\
\hline
\multirow{2}{*}{\textbf{Navigation Pos. Error (mm)}} 
  & SIL & 6.731 & 12.384 & 8.721 $\pm$ 3.162  \\
  & HIL & 5.050 & 6.091 & 8.367 $\pm$ 2.679 \\
\hline
\multirow{2}{*}{\textbf{Navigation Vel. Error (mm/s)}} 
  & SIL & 0.047 & 0.015 & 0.024 $\pm$ 0.011 \\
  & HIL & 0.031 & 0.006 & 0.021 $\pm$ 0.009 \\
\hline
\multirow{2}{*}{\textbf{Control Pos. Error (m)}} 
  & SIL & 0.588 & 0.851 & \multicolumn{1}{c}{-} \\
  & HIL & 0.364 & 0.715 & \multicolumn{1}{c}{-} \\
\hline
\end{tabular}
\label{tab:close-range-results-dgtl}
\end{table}

\begin{figure}[htb]
    \centering
    \includegraphics[width=0.75\linewidth]{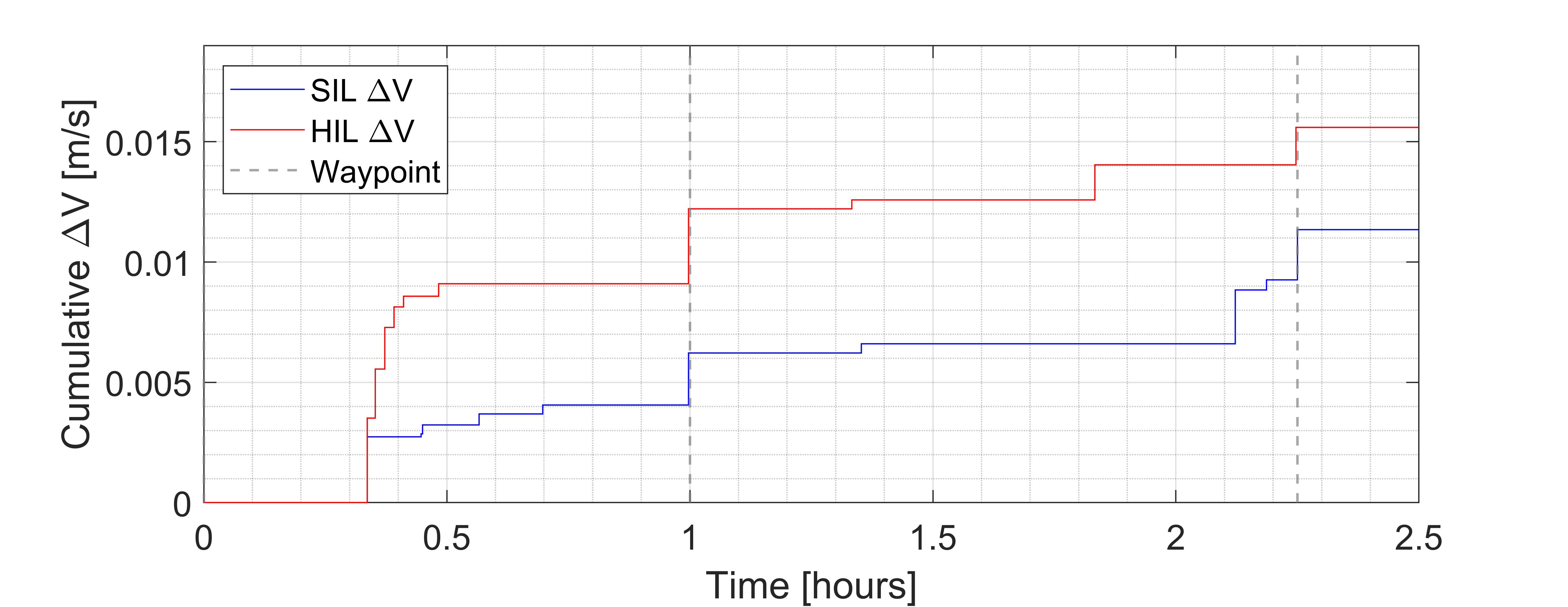}
    \caption{SIL and HIL cumulative $\Delta V$ in Experiment \#3. In total, SIL executed 12 maneuvers (\SI{12.476} {mm/\second}) and HIL executed 11 maneuvers (\SI{15.596} {mm/\second}). (25\% increase from HIL over SIL). The dashed vertical lines indicate the time associated with each waypoint in the trajectory.}
    \label{fig:dgtl-dv-cumulative}
\end{figure}

The resulting maneuver plans, and hence cumulative $\Delta V$ profiles, however, are not identical, as seen in \autoref{fig:dgtl-dv-cumulative}.
The maneuver plan is sensitive even to small variations in state errors, which reflect differences between the emulated (SIL) and true (HIL) distributions of the measurement noise sources, especially prior to steady state convergence as aforementioned.
The downstream effects of these differences in the HIL and SIL simulations are apparent in Figure \ref{fig:dgtl-dv-cumulative}, where under HIL conditions, additional $\Delta V$ is expended during the transients in response to higher state errors. 
This led to large but distinct initial maneuvers between $t = 20$ to $30$ minutes, followed by a maneuver profile thereafter that is more consistent between both SIL and HIL, since the filters exhibit similar steady-state characteristics.

\section{Conclusion}
This work presents an integrated, hybrid digital and robotic twinning framework for end-to-end verification and validation (V\&V) of autonomous GNC stacks for Rendezvous, Proximity Operations (RPO) and Formation Flying (FF) applications. 
The framework combines Software-in-the-Loop (SIL) and Hardware-in-the-Loop (HIL) methodologies to evaluate the performance of the GNC stack under real-world conditions.
By incorporating three testbeds at Stanford's Space Rendezvous Lab (SLAB), the proposed pipeline supports closed-loop testing across full mission scenarios with real spacecraft hardware. 
The architecture enables validation of multi-modal sensing (vision- and RF-based) and control strategies, accommodates cooperative and uncooperative targets, and captures transitions between sensing regimes. 

The framework was validated through the execution of a full-range RPO mission spanning \SI{75}{\kilo\meter} to \SI{7}{\meter} using both vision-based and RF sensing modes. 
The RPO mission was divided into three experiments that directly compare SIL and HIL to quantify the sim-to-world gap.
The results help to assess the robustness of navigation algorithms to sensor imperfections and measurement noise, evaluate the fidelity of digital twins, and predict mission-relevant metrics such as cumulative $\Delta V$ and control error.
Ultimately, the experiments demonstrate the framework's ability to broadly reproduce comparable scenario realizations between SIL and HIL experiments.
At the same time, the hardware-based testbeds sometimes provided more realistic and challenging measurements to the GNC stack, allowing operators to understand flight software's robustness to unmodeled effects.
Overall, this modular, extensible approach advances the state-of-the-art in GNC V\&V by enabling high-fidelity, scalable assessments of autonomy stacks under realistic operational conditions. 

\subsubsection{Lessons Learned} 

While HIL is time-consuming, it reveals valuable failure modes that are not uncovered through SIL campaigns alone. A key lesson learned was the importance of deliberate and incremental experiment planning. Given the modularity of the GNC stack, individual modules should first be isolated and evaluated in SIL to establish baseline performance. Once this baseline is set, additional software and hardware elements can be incrementally introduced to ensure successful integration. When problematic interactions arise, they are easier to diagnose and resolve because each component has already been independently benchmarked.
Additionally, the determinism of the simulation environment is critical to the testing process, as it allows bugs and unexpected behaviors to be reliably reproduced and diagnosed. This, in turn, enables system designers to refine algorithms to meet mission requirements.

Future work will include extending the fidelity of attitude propagation, further validation of digital and robotic twins against space-based data, improved photometric modeling and calibration of the OpenGL-based renderer, sensor fusion between diverse measurement types, and integration of a representative flight computer to host flight software under test. Looking forward, the presented digital and robotic twinning framework is well suited for the validation of precise formation flying missions, such as the NASA-funded STarlight Acquisition and Reflection toward Interferometry (STARI) mission\cite{RizzaSpaceIF}.

\section{Acknowledgments}
This work is partially funded by Infinite Orbits through AFWERX contract FA864924P1021. 
The authors would like to thank Justin Kruger and Hugo Dage for their support and insightful discussions. The authors would also like to thank Gregory Zin for his support with the hardware interfaces. 
Finally, the authors wish to thank Juergen Bosse for his timely assistance with the TRON testbed.


\clearpage
\bibliographystyle{AAS_publication}   
\bibliography{references}   

\begin{thebibliography}{10}

\bibitem{sharma_pose_2019}
S.~Sharma and S.~D’Amico, ``Pose Estimation for Non-Cooperative Spacecraft Rendezvous Using Neural Networks,''  {\em 2019 AAS/AIAA Astrodynamics Specialist Conference, Ka’anapali, Maui, HI}, January 13-17 2019.

\bibitem{park2021tron}
T.~H. Park, J.~Bosse, and S.~D'Amico, ``Robotic Testbed for Rendezvous and Optical Navigation: Multi-Source Calibration and Machine Learning Use Cases,''  {\em 2021 AAS/AIAA Astrodynamics Specialist Conference, Big Sky, Virtual}, August 9-11 2021.

\bibitem{beierle_design_2017}
C.~Beierle, J.~Sullivan, and S.~D’Amico, ``Design and Utilization of the Stanford Vision-Based Navigation Testbed for Spacecraft Rendezvous,''  {\em Proceedings of the 9th International Workshop on Satellite Constellations and Formation Flying, University of Colorado}, June 19-21 2017.

\bibitem{beierle2019high}
C.~R. Beierle, {\em High fidelity validation of vision-based sensors and algorithms for spaceborne navigation}.
\newblock PhD thesis, Stanford University, 2019.

\bibitem{giralo2018testbed}
V.~Giralo and S.~D'Amico, ``Development of the Stanford GNSS Navigation Testbed for Distributed Space Systems,''  {\em Proceedings of the 2018 International Technical Meeting of the Institute of Navigation}, January 2018, pp.~837--856, 10.33012/2018.15544.

\bibitem{ansys_stk}
{Analytical Graphics, Inc. (AGI)}, ``{ANSYS Systems Tool Kit (STK)},''  \url{https://www.agi.com/stk}, 2025.

\bibitem{hughes2016general}
{NASA Goddard Space Flight Center}, ``General Mission Analysis Tool (GMAT), Version R2022a,''  \url{https://software.nasa.gov/software/GSC-19097-1}, 2022.
\newblock NASA Software Catalog, GSC-19097-1.

\bibitem{freeflyer}
{A.I. Solutions, Inc.}, ``{FreeFlyer},''  \url{https://ai-solutions.com}, 2025.

\bibitem{kenneally_basilisk_2020}
P.~W. Kenneally, S.~Piggott, and H.~Schaub, ``Basilisk: {A} {Flexible}, {Scalable} and {Modular} {Astrodynamics} {Simulation} {Framework},''  {\em Journal of Aerospace Information Systems}, Vol.~17, Sept. 2020, pp.~496--507.
\newblock Publisher: American Institute of Aeronautics and Astronautics (AIAA), 10.2514/1.i010762.

\bibitem{aerospace_toolbox}
{The MathWorks Inc.}, ``Aerospace Toolbox, Version 24.2 (R2024b),''  \url{https://www.mathworks.com/products/aerospace-toolbox.html}, 2024.

\bibitem{choo2004scibox}
T.~H. Choo and J.~P. Skura, ``SciBox: a software library for rapid development of science operation simulation, planning, and command tools,''  {\em Johns Hopkins APL Technical Digest}, Vol.~25, No.~2, 2004, pp.~154--162.

\bibitem{garcia2025dshell}
J.~Garcia-Bonilla, C.~Leake, A.~Elmquist, T.~D. Hasseler, V.~Steyert, A.~Gaut, and A.~Jain, ``Dshell-DARTS: A Reusability-Focused Multi-Mission Aerospace and Robotics Simulation Toolkit,''  {\em 2025 IEEE Aerospace Conference}, 2025, pp.~1--13, 10.1109/AERO63441.2025.11068690.

\bibitem{penn2016trick}
J.~Penn and A.~Lin, ``The {Trick} {Simulation} {Toolkit}: {A} {NASA}/{Opensource} {Framework} for {Running} {Time} {Based} {Physics} {Models},''  {\em {AIAA} {Modeling} and {Simulation} {Technologies} {Conference}}, 2016, 10.2514/6.2016-1187.

\bibitem{cols-margenet_modular_2016}
M.~Cols-Margenet, H.~Schaub, and S.~Piggott, ``Modular {Attitude} {Guidance} {Development} using the {Basilisk} {Software} {Framework},''  {\em {AIAA} {SPACE} 2016}, Long Beach, California, American Institute of Aeronautics and Astronautics, Sept. 2016, 10.2514/6.2016-5538.

\bibitem{basiliskscheduling}
H.~Schaub, ``Execution of Basilisk Modules,''  \url{https://hanspeterschaub.info/basilisk/Learn/bskPrinciples/bskPrinciples-2a.html}, 2024.
\newblock Basilisk 2.4.0 Documentation.

\bibitem{tbell2025sim}
T.~Bell and S.~D'Amico, ``Event-Driven Simulation for Rapid Iterative Development of Distributed Space Flight Software,''  {\em 2025 IEEE Aerospace Conference}, 2025, pp.~1--19, 10.1109/AERO63441.2025.11068699.

\bibitem{guffanti2023autonomous}
T.~Guffanti, T.~Bell, S.~Y.~W. Low, M.~Murray-Cooper, and S.~D'Amico, ``Autonomous Guidance Navigation and Control of the VISORS Formation-Flying Mission,''  {\em 2023 AAS/AIAA Astrodynamics Specialist Conference, Big Sky, Montana}, 2023.

\bibitem{Kruger2024}
J.~J. Kruger, T.~Guffanti, T.~H. Park, M.~Murray-Cooper, S.~Y. Low, T.~Bell, S.~D'Amico, C.~W. Roscoe, and J.~Westphal, ``Adaptive End-to-End Architecture for Autonomous Spacecraft Navigation and Control During Rendezvous and Proximity Operations,''  {\em AIAA SciTech Forum}, 2024, 10.2514/6.2024-0430.

\bibitem{wilde_historical_2019}
M.~Wilde, C.~Clark, and M.~Romano, ``Historical survey of kinematic and dynamic spacecraft simulators for laboratory experimentation of on-orbit proximity maneuvers,''  {\em Progress in Aerospace Sciences}, Vol.~110, Oct. 2019, p.~100552, 10.1016/j.paerosci.2019.100552.

\bibitem{leitner2001gnsstestbed}
J.~Leitner, ``A hardware-in-the-loop testbed for spacecraft formation flying applications,''  {\em 2001 IEEE Aerospace Conference Proceedings (Cat. No.01TH8542)}, Vol.~2, 2001, pp.~2/615--2/620 vol.2, 10.1109/AERO.2001.931240.

\bibitem{burns2004gnsstestbed}
R.~Burns, B.~Naasz, D.~Gaylor, and J.~Higinbotham, ``An {Environment} for {Hardware}-in-the-{Loop} {Formation} {Navigation} and {Control},''  {\em {AIAA}/{AAS} {Astrodynamics} {Specialist} {Conference} and {Exhibit}}.
\newblock \_eprint: https://arc.aiaa.org/doi/pdf/10.2514/6.2004-4735, 10.2514/6.2004-4735.

\bibitem{gaias2012gnsstestbed}
G.~Gaias, J.-S. Ardaens, and S.~D'Amico, ``{Formation flying testbed at DLR's German Space Operations Center (GSOC)},''  {\em 8th International ESA Conference on GNC (GNC 2011)}, 2012.

\bibitem{fedora2008spirent}
N.~Fedora, C.~Ford, and P.~Boulton, ``A Versatile Solution for Testing GPS/Inertial Navigation Systems,''  {\em Proceedings of the 21st International Technical Meeting of the Satellite Division of The Institute of Navigation (ION GNSS 2008)}, 2008, pp.~1227--1236.

\bibitem{heinrichs2007ifen}
G.~Heinrichs, M.~Irsigler, R.~Wolf, J.~Winkel, and G.~Prokoph, ``NavX (R)-NCS-The First Galileo/GPS Full RF Navigation Constellation Simulator,''  {\em Proceedings of the 20th International Technical Meeting of the Satellite Division of The Institute of Navigation (ION GNSS 2007)}, 2007, pp.~1323--1328.

\bibitem{biswas2014gnsstestbed}
S.~Biswas, L.~Qiao, and A.~Dempster, ``Space-borne GNSS based orbit determination using a SPIRENT GNSS simulator,''  {\em 15th Australian Space Research Conference, Adelaide, Australia}, 2014.

\bibitem{peng2017gnsstestbed}
Y.~Peng, {\em GNSS-based Spacecraft Formation Flying Simulation and Ionospheric Remote Sensing Applications}.
\newblock PhD thesis, Virginia Tech, 2017.

\bibitem{connor_os}
C.~Beierle and S.~D’Amico, ``Variable-Magnification Optical Stimulator for Training and Validation of Spaceborne Vision-based Navigation,''  {\em Journal of Spacecraft and Rockets}, Vol.~56, No.~4, 2019, pp.~1060--1072, 10.2514/1.A34337.

\bibitem{CEL}
B.~Boone, J.~Bruzzi, W.~Dellinger, B.~Kluga, and K.~Strobehn, ``Optical simulator and testbed for spacecraft star tracker development,''  {\em Optical Modeling and Performance Predictions II}, Vol.~5867, SPIE, 2005, pp.~318--331, 10.1117/12.619133.

\bibitem{OSI}
M.~A. Samaan, S.~R. Steffes, and S.~Theil, ``Star tracker real-time hardware in the loop testing using optical star simulator,''  {\em Spaceflight Mechanics}, Vol.~140, 2011.

\bibitem{roessler_optical_2014}
D.~Roessler, D.~A. Pedersen, M.~Benn, and J.~L. Jørgensen, ``Optical stimulator for vision-based sensors,''  {\em Advanced Optical Technologies}, Vol.~3, Apr. 2014, pp.~199--207, 10.1515/aot-2013-0045.

\bibitem{rufino_real-time_2013}
G.~Rufino, D.~Accardo, M.~Grassi, G.~Fasano, A.~Renga, and U.~Tancredi, ``Real-{Time} {Hardware}-in-the-{Loop} {Tests} of {Star} {Tracker} {Algorithms},''  {\em International Journal of Aerospace Engineering}, Vol.~2013, No.~1, 2013, p.~505720, 10.1155/2013/505720.

\bibitem{boge2012epos}
T.~Boge, H.~Benninghoff, M.~Zebenay, and F.~Rems, ``Using robots for advanced rendezvous and docking simulation,''  {\em Proc. Simulation and EGSE facilites for Space Programmes (SESP), Noordwijk, The Netherlands}, 2012.

\bibitem{boge_new_2002}
T.~Boge and E.~Schreutelkamp, ``A New Commanding and Control Environment for Rendezvous and Docking Simulations at the EPOS-Facility,''  {\em 7th International Workshop on Simulation for European Space Programmes}, 2002, pp.~215--222.

\bibitem{mietner_european_2017}
C.~Mietner, ``European {Proximity} {Operations} {Simulator} 2.0 ({EPOS}) - {A} {Robotic}-{Based} {Rendezvous} and {Docking} {Simulator},''  {\em Journal of Large-Scale Research Facilities JLSRF}, Vol.~3, Apr. 2017, 10.17815/jlsrf-3-155.

\bibitem{milenkovic2012lockheed}
Z.~Milenkovic and C.~D'Souza, ``The Space Operations Simulation Center (SOSC) and closed-loop hardware testing for Orion rendezvous system design,''  {\em AIAA Guidance, Navigation, and Control Conference}, 2012, p.~5034.

\bibitem{kruger_tron_2010}
H.~Krüger and S.~Theil, ``{TRON} - {Hardware}-in-the-{Loop} {Test} {Facility} for {Lunar} {Descent} and {Landing} {Optical} {Navigation},''  {\em IFAC Proceedings Volumes}, Vol.~43, Jan. 2010, pp.~265--270, 10.3182/20100906-5-JP-2022.00046.

\bibitem{j_d_mitchell_automated_2007}
J.~D. Mitchell, S.~P. Cryan, D.~Strack, L.~L. Brewster, M.~J. Williamson, R.~T. Howard, and A.~S. Johnston, ``Automated Rendezvous and Docking Sensor Testing at the Flight Robotics Laboratory,''  {\em 2007 IEEE Aerospace Conference}, 2007, pp.~1--16, 10.1109/AERO.2007.352723.

\bibitem{POSEIDYN}
M.~Wilde, S.~T.~K. Choon, and M.~Romano, ``Kinematic and Dynamic Spacecraft Maneuver Simulators for Verification and Validation of Space Robotic Systems,''  {\em AIAA Scitech 2020 Forum}, 2020, 10.2514/6.2020-1919.

\bibitem{sternberg2018jet}
D.~C. Sternberg, C.~Pong, N.~Filipe, S.~Mohan, S.~Johnson, and L.~Jones-Wilson, ``Jet Propulsion Laboratory small satellite dynamics testbed simulation: On-orbit performance model validation,''  {\em Journal of Spacecraft and Rockets}, Vol.~55, No.~2, 2018, pp.~322--334, 10.2514/1.A33806.

\bibitem{jpl_newer}
J.~D. Wapman, D.~C. Sternberg, K.~Lo, M.~Wang, L.~Jones-Wilson, and S.~Mohan, ``Jet Propulsion Laboratory Small Satellite Dynamics Testbed Planar Air-Bearing Propulsion System Characterization,''  {\em Journal of Spacecraft and Rockets}, Vol.~58, No.~4, 2021, pp.~954--971, 10.2514/1.A34857.

\bibitem{tappe_development_2009}
J.~A. Tappe, {\em Development of {Star} {Tracker} {System} for {Accurate} {Estimation} of {Spacecraft} {Attitude}}.
\newblock PhD thesis, Naval Postgraduate School, 2009.

\bibitem{mote_collision-inclusive_2020}
M.~Mote, M.~Egerstedt, E.~Feron, A.~Bylard, and M.~Pavone, ``Collision-{Inclusive} {Trajectory} {Optimization} for {Free}-{Flying} {Spacecraft},''  {\em Journal of Guidance, Control, and Dynamics}, Mar. 2020.
\newblock Publisher: American Institute of Aeronautics and Astronautics, 10.2514/1.G004788.

\bibitem{GTechAstros}
P.~Tsiotras, ``ASTROS: A 5DOF experimental facility for research in space proximity operations,''  {\em Advances in the Astronautical Sciences}, Vol.~151, 01 2014, pp.~717--730.

\bibitem{nakka_six_2018}
Y.~K. Nakka, R.~C. Foust, E.~S. Lupu, D.~B. Elliott, I.~S. Crowell, S.-J. Chung, and F.~Y. Hadaegh, ``Six {Degree}-of-{Freedom} {Spacecraft} {Dynamics} {Simulator} for {Formation} {Control} {Research},''  {\em 2018 AAS/AIAA Astrodynamics Specialist Conference}, Aug. 2018, pp.~1--20.

\bibitem{dormand_family_1980}
J.~R. Dormand and P.~J. Prince, ``A family of embedded {Runge}-{Kutta} formulae,''  {\em Journal of Computational and Applied Mathematics}, Vol.~6, Mar. 1980, pp.~19--26, 10.1016/0771-050X(80)90013-3.

\bibitem{picone_nrlmsise-00_2002}
J.~M. Picone, A.~E. Hedin, D.~P. Drob, and A.~C. Aikin, ``{NRLMSISE}-00 empirical model of the atmosphere: {Statistical} comparisons and scientific issues,''  {\em Journal of Geophysical Research: Space Physics}, Vol.~107, No.~A12, 2002, pp.~SIA 15--1--SIA 15--16, 10.1029/2002JA009430.

\bibitem{novatelmanual}
Novatel, ``OEM7 Commands and Logs Reference Manual,''  2025.

\bibitem{garrido2014automatic}
S.~Garrido-Jurado, R.~Mu{\~n}oz-Salinas, F.~J. Madrid-Cuevas, and M.~J. Mar{\'\i}n-Jim{\'e}nez, ``Automatic generation and detection of highly reliable fiducial markers under occlusion,''  {\em Pattern Recognition}, Vol.~47, No.~6, 2014, pp.~2280--2292, 10.1016/j.patcog.2014.01.005.

\bibitem{tabb_solving_rwhe}
A.~Tabb and K.~Ahmad~Yousef, ``Solving the robot-world hand-eye(s) calibration problem with iterative methods,''  {\em Machine Vision and Applications}, 08 2017, pp.~1--22, 10.1007/s00138-017-0841-7.

\bibitem{d2012prisma}
S.~D’Amico, P.~Bodin, M.~Delpech, and R.~Noteborn, ``PRISMA,''  {\em Distributed space missions for earth system monitoring}, pp.~599--637, Springer, 2013, 10.1007/978-1-4614-4541-8.

\bibitem{teunissen1994lambda}
P.~Teunissen, ``A new method for fast carrier phase ambiguity estimation,''  {\em Proceedings of 1994 IEEE Position, Location and Navigation Symposium-PLANS'94}, IEEE, 1994, pp.~562--573.

\bibitem{low2024coupling}
S.~Y. Low and S.~D’Amico, ``Precise distributed satellite navigation: Differential GPS with sensor-coupling for integer ambiguity resolution,''  {\em 2024 IEEE Aerospace Conference}, IEEE, 2024, pp.~1--18.

\bibitem{low2026digital}
S.~Y. Low, T.~Bell, and S.~D’Amico, ``Flight-Ready Precise and Robust Carrier-Phase GNSS Navigation Software for Distributed Space Systems,''  {\em AIAA Scitech 2026 Forum}, 2026.

\bibitem{kruger2021autonomous}
J.~Kruger and S.~D’Amico, ``Autonomous angles-only multitarget tracking for spacecraft swarms,''  {\em Acta Astronautica}, Vol.~189, 2021, pp.~514--529, https://doi.org/10.1016/j.actaastro.2021.08.049.

\bibitem{park_robust_2023}
T.~H. Park and S.~D’Amico, ``Robust multi-task learning and online refinement for spacecraft pose estimation across domain gap,''  {\em Advances in Space Research}, Mar. 2023, 10.1016/j.asr.2023.03.036.

\bibitem{giralo2019digital}
V.~Giralo and S.~D’Amico, ``Distributed multi-{GNSS} timing and localization for nanosatellites,''  {\em NAVIGATION}, Vol.~66, No.~4, 2019, pp.~729--746, 10.1002/navi.337.

\bibitem{sullivan_generalized_2021}
J.~Sullivan, A.~W. Koenig, J.~Kruger, and S.~D’Amico, ``Generalized {Angles}-{Only} {Navigation} {Architecture} for {Autonomous} {Distributed} {Space} {Systems},''  {\em Journal of Guidance, Control, and Dynamics}, Vol.~44, June 2021, pp.~1087--1105, 10.2514/1.G005439.

\bibitem{park_adaptive_2023}
T.~H. Park and S.~D’Amico, ``Adaptive {Neural}-{Network}-{Based} {Unscented} {Kalman} {Filter} for {Robust} {Pose} {Tracking} of {Noncooperative} {Spacecraft},''  {\em Journal of Guidance, Control, and Dynamics}, Vol.~46, Sept. 2023, pp.~1671--1688, 10.2514/1.G007387.

\bibitem{chang2005mlambda}
X.-W. Chang, X.~Yang, and T.~Zhou, ``MLAMBDA: a modified LAMBDA method for integer least-squares estimation,''  {\em Journal of Geodesy}, Vol.~79, 01 2005, pp.~552--565, 10.1007/s00190-005-0004-x.

\bibitem{chernick_closed-form_2021}
M.~Chernick and S.~D’Amico, ``Closed-{Form} {Optimal} {Impulsive} {Control} of {Spacecraft} {Formations} {Using} {Reachable} {Set} {Theory},''  {\em Journal of Guidance, Control, and Dynamics}, Vol.~44, Jan. 2021, pp.~25--44.
\newblock Publisher: American Institute of Aeronautics and Astronautics, 10.2514/1.G005218.

\bibitem{chernick_new_2018}
M.~Chernick and S.~D’Amico, ``New {Closed}-{Form} {Solutions} for {Optimal} {Impulsive} {Control} of {Spacecraft} {Relative} {Motion},''  {\em Journal of Guidance, Control, and Dynamics}, Vol.~41, Feb. 2018, pp.~301--319, 10.2514/1.G002848.

\bibitem{koenig2020fast}
A.~W. Koenig and S.~D'Amico, ``Fast algorithm for fuel-optimal impulsive control of linear systems with time-varying cost,''  {\em IEEE Transactions on Automatic Control}, Vol.~66, No.~9, 2020, pp.~4029--4042.

\bibitem{damico_noncooperative_2013}
S.~D’Amico, J.-S. Ardaens, G.~Gaias, H.~Benninghoff, B.~Schlepp, and J.~L. Jørgensen, ``Noncooperative {Rendezvous} {Using} {Angles}-{Only} {Optical} {Navigation}: {System} {Design} and {Flight} {Results},''  {\em Journal of Guidance, Control, and Dynamics}, Vol.~36, Nov. 2013, pp.~1576--1595, 10.2514/1.59236.

\bibitem{damico}
S.~D'Amico, {\em Autonomous formation flying in low earth orbit}.
\newblock PhD thesis, TU Delft, 2010.

\bibitem{d2006proximity}
S.~D'Amico and O.~Montenbruck, ``Proximity operations of formation-flying spacecraft using an eccentricity/inclination vector separation,''  {\em Journal of Guidance, Control, and Dynamics}, Vol.~29, No.~3, 2006, pp.~554--563, 10.2514/1.15114.

\bibitem{RizzaSpaceIF}
A.~Rizza, E.~Foss, J.~Monnier, J.~Cutler, and S.~D'Amico, ``Space Interferometry Formation Design using Relative Orbital Elements: the STARI Mission,''  {\em 2026 IEEE Aerospace Conference}.

\bibitem{psiaki2007modeling}
M.~L. Psiaki and S.~Mohiuddin, ``Modeling, analysis, and simulation of GPS carrier phase for spacecraft relative navigation,''  {\em Journal of Guidance, Control, and Dynamics}, Vol.~30, No.~6, 2007, pp.~1628--1639.

\end{thebibliography}

\clearpage 
\section{Appendix A}
\label{sec:appendix_1}



\autoref{tab:gnss-emu-sig} tabulates the GNSS measurement noise and measurement visibility modeling, while \autoref{tab:gnss-emu-sc} tabulates relevant body-frame component uncertainties that impact GNSS measurements.
Both SIL and HIL measurement generation environments adopt the exact same parameters.

\begin{table}[H]
\centering
\begin{minipage}[t]{0.48\textwidth}
\caption{Default GNSS signal-in-space model}
\label{tab:gnss-emu-sig}
\centering
\footnotesize
\renewcommand{\arraystretch}{1.1}
\begin{tabular}{>{\bfseries}p{2.5cm} p{3.5cm}}
\toprule
GNSS signals & GPS L1/L2, GAL E1, BDS B1I \\
GNSS aperture & 170 degrees \\
GNSS gain pattern & Spherical only \\
GNSS ephemeris & Uncorrected broadcast \\
Ionospheric model & Klobuchar (delay) \\
Earth horizon mask & -20$^\circ$ from horizon \\
Receiver mask & 190 degrees \\
Simulated LNA & 33 dBW \\
Gain pattern & ANTCOM 1.9G1215A \\
Thermal noise model & Psiaki et al \cite{psiaki2007modeling} \\
\bottomrule
\end{tabular}
\end{minipage}
\hfill
\begin{minipage}[t]{0.48\textwidth}
\caption{Default body-component uncertainty}
\label{tab:gnss-emu-sc}
\centering
\footnotesize
\renewcommand{\arraystretch}{1.1}
\begin{tabular}{>{\bfseries}p{3.5cm} c c c}
\toprule
Parameter & Unit & Dist. & Value \\
\midrule
Spacecraft mass & kg & Gaussian & 0.04 \\
Spacecraft COM & mm & Uniform & 3 \\
GNSS antenna PCO bias & mm & -- & 5 \\
GNSS antenna PCO noise & mm & Gaussian & 1 \\
GNSS antenna direction & arcsec & Gaussian & 30 \\
Attitude knowledge & arcsec & Gaussian & 20 \\
\bottomrule
\end{tabular}
\end{minipage}
\end{table}
\newpage
\section{Appendix B}
\label{sec:appendix_2}

\subsection{Additional Results: Experiment 1}

\subsection{Additional Results: Experiment 2}

\begin{figure}[h]
    \centering
    \includegraphics[width=0.9\linewidth]{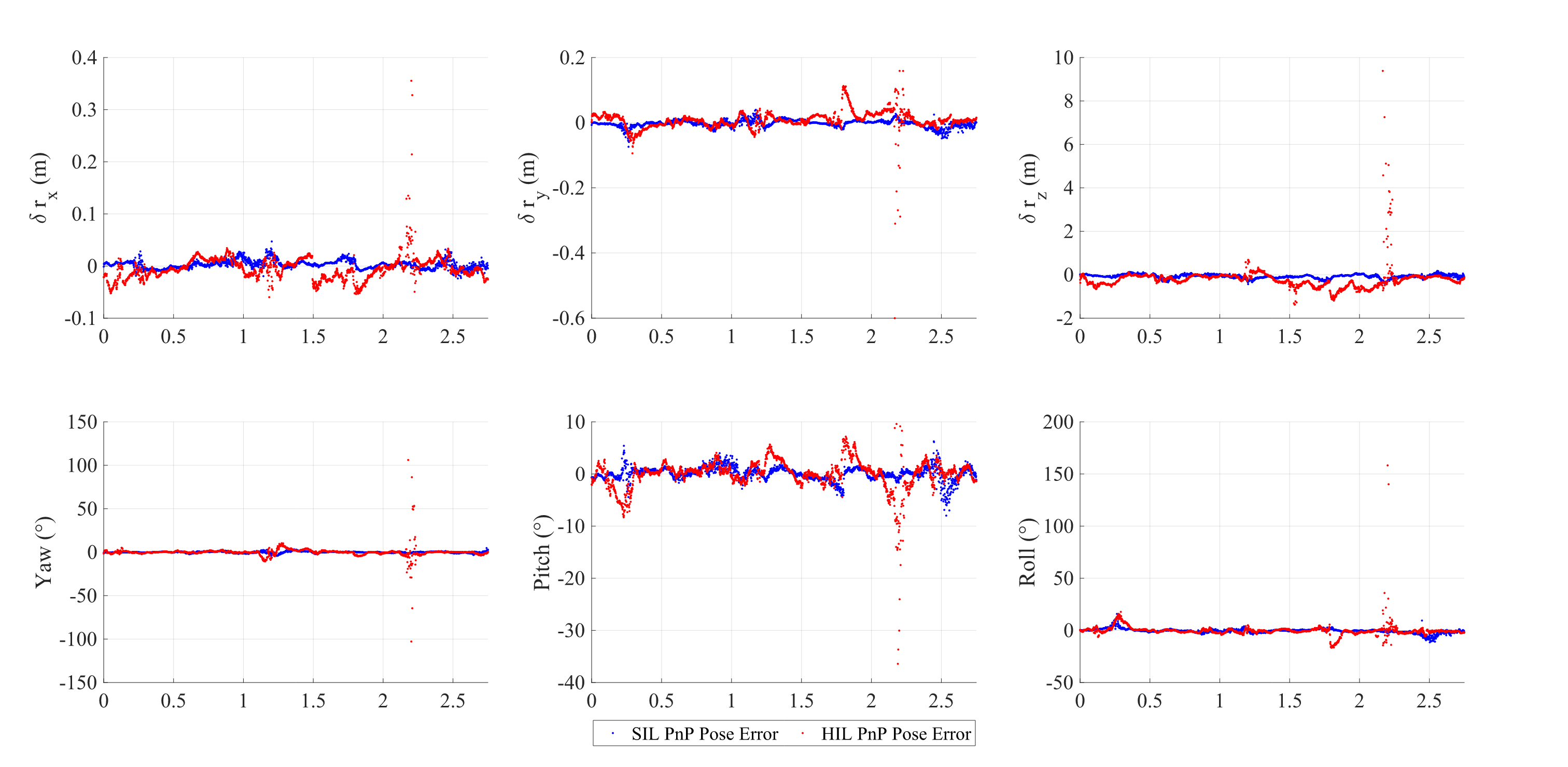}
    \caption{Navigation performance of Experiment \#2: PnP pose error of SPN measured as the difference between the PnP solution from the detected features in the image and the true pose. Errors shown for both SIL and HIL experiments, and given in the camera frame. PnP pose errors are defined as the error between the PnP pose solution using SPN's detected keypoints and the true pose from the simulation.}
    \label{fig:spn_results_spn_measurement_errors}
\end{figure}







\end{document}